\documentclass{article}

\usepackage[preprint]{neurips_2025}

\usepackage[utf8]{inputenc} 
\usepackage[T1]{fontenc}    
\usepackage{hyperref}       
\usepackage{url}            
\usepackage{booktabs}       
\usepackage{amsfonts}       
\usepackage{caption}        
\usepackage{graphicx}       
\usepackage{nicefrac}       
\usepackage{microtype}      
\usepackage{xcolor}         
\usepackage{amsmath}
\usepackage{algorithmic}
\usepackage{algorithm}
\usepackage{pifont}
\usepackage{wrapfig}
\usepackage{booktabs}
\PassOptionsToPackage{numbers,compress}{natbib}

\title{Sparse Reasoning is Enough: Biological-Inspired Framework for Video Anomaly Detection with Large Pre-trained Models}

\author{%
  He Huang\thanks{These authors contributed equally to this work.}\\
  School of Computer Science\\
  Peking University\\
  \textit{binarystars2077@gmail.com}
  \And 
  Zixuan Hu\footnotemark[1]\\
  School of Computer Science\\
  Peking University\\
  \textit{hzxuan@pku.edu.cn}
  \And
  Dongxiao Li \\
  School of Computer Science\\
  Peking University\\
  \textit{llldddxxx03@gmail.com}
  \And
  Yao Xiao \\
  Fuzhou Chengtou New Infrastructure Co., Ltd\\
  \textit{yx2513@nyu.edu}
  \And
  Ling-Yu Duan\thanks{Corresponding author. Email: lingyu@pku.edu.cn} \\
  School of Computer Science\\
  Peking University\\
  \textit{lingyu@pku.edu.cn}
}

\begin{document}

\maketitle

\begin{abstract}
Video anomaly detection (VAD) plays a vital role in real-world applications such as security surveillance, autonomous driving, and industrial monitoring. Recent advances in large pre-trained models have opened new opportunities for training-free VAD by leveraging rich prior knowledge and general reasoning capabilities. However, existing studies typically rely on dense frame-level inference, incurring high computational costs and latency. This raises a fundamental question: Is this dense reasoning truly necessary when using powerful pre-trained models in VAD systems? To answer this, we propose \textbf{ReCoVAD}, a novel framework inspired by the human nervous system’s dual \textbf{re}flex and \textbf{co}nscious pathways, enabling selective frame processing to reduce redundant computation. ReCoVAD consists of two core pathways: i) The Reflex pathway, a lightweight CLIP-based component, fuses visual features with prototype prompts to produce decision vectors. These are used to query a dynamic memory of past frames and their anomaly scores, enabling the system to rapidly determine whether to respond immediately or escalate the frame for further reasoning. ii) The Conscious pathway, a medium-scale vision-language model, generates textual event descriptions and refined anomaly scores for novel frames. It continuously updates the memory and prototype prompts, while an integrated LLM periodically reviews accumulated descriptions to identify unseen anomalies, correct errors, and refine prototypes. Our extensive experiments show that ReCoVAD reaches state-of-the-art training-free performance while processing only \textbf{28.55\%}/\textbf{16.04\%} of frames used by the previous methods in UCF-Crime/XD-Violence datasets, demonstrating that sparse reasoning is enough for effective large-model-based VAD.
\end{abstract}

\section{Introduction}
Video Anomaly Detection (VAD) aims to automatically identify anomalous events in video streams that exhibit deviation from normal patterns. It has attracted substantial attention due to its wide-range applications, including security surveillance \cite{liu2018future,sultani2018real}, autonomous driving \cite{yao2020and,hu2025adaptive}, and industrial monitoring \cite{pang2021deep}, etc. Due to limited data and computational resources in the edge devices (e.g., camera, vehicle), it’s crucial to develop efficient and generalizable VAD systems to enable in-time responses to diverse anomalies.
\begin{figure}[t]
    \centering
     \includegraphics[width=1.0\linewidth]
     {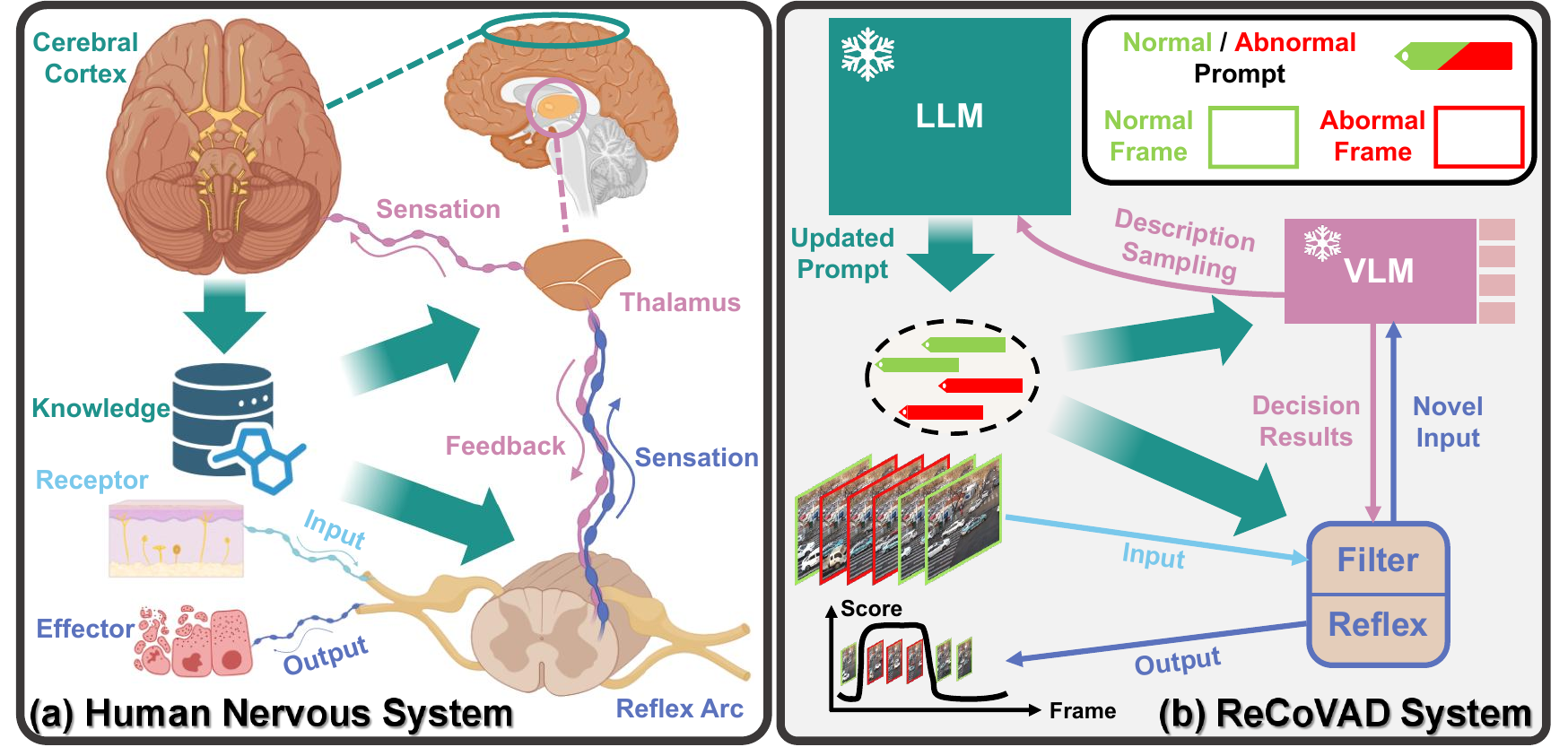}
     \caption{(a) The nervous system routes familiar signals through low-cost reflex arcs, while allocating cortical resources only to novel or complex inputs via the thalamus–cortex loop. The Bidirectional feedback enables top-down modulation of reflexes and bottom-up filtering of redundant information (b) ReCoVAD mirrors this architecture: the reflex pathway filters redundant frames by leveraging visual signals and prompt-conditioned decision vectors, while the conscious pathway applies VLM/LLM-based reasoning for novel inputs. Through a feedback loop, it refines both memory and prompts, progressively enhancing selectivity.}
     \label{fig:teaser}
\end{figure}

While conventional VAD models exhibit high efficiency in limited conditions, they suffer from poor generalization and remain fragile in dynamic real-world deployments \cite{openvad,liu2023generalized,zanella2024harnessing,wu2024open,hu2025beyond}.
To overcome this, recent works\cite{zanella2024harnessing,yang2024follow,kim2023unsupervised,hu2024lead,tang2024hawk,du2024uncovering} explore the integration of Large Visual Language Model (LVLM) and Large Language Model (LLM), leveraging their broad prior knowledge from the large-scale pre-training.
This emerging paradigm involves using i) LVLMs to produce rich textual descriptions of video frames and ii) LLMs to infer anomalous events from these descriptions \cite{zanella2024harnessing}.
By decoupling anomaly detection from low-level visual cues, this system achieves significantly improved generalization.

Despite these advantages, large-model VAD systems face two critical deployment challenges: i) High computational cost of inference \cite{samsi2023words,zhou2024survey}, and ii) Dense frame-level analysis to ensure anomaly coverage.
These challenges inevitably compound the overall computational burden.
This raises a fundamental question: When leveraging powerful foundation models, is frequent inference across densely sampled frames truly necessary?
While prior studies have validated the benefits of dense coverage in traditional VAD pipelines \cite{zanella2024harnessing,yang2024follow,wu2024open}, its utility in large-model systems remains underexplored and potentially suboptimal.

To address this challenge, we draw inspiration from how the human nervous system reduces the brain's burden through hierarchical processing \cite{Meunier_2009}, allocating cognitive resources selectively. As shown in Figure \ref{fig:teaser}(a), it employs two complementary processing pathways: i) \textit{Reflex} pathway that rapidly filters and responds to familiar, redundant signals via reflex arcs and ii) \textit{Conscious} pathway where the thalamus performs primary analysis \cite{dehghani2019computational} and cortical regions engage in further reasoning only on novel or complex inputs \cite{nani2019neural}. These pathways interact through a bidirectional feedback loop \cite{zagha2020shaping}: the conscious pathway influences reflex responses through memory and knowledge, while the reflex filters out redundant signals, reducing the cognitive load for the conscious pathway. This paradigm suggests that dense and uniform reasoning is not always necessary, inspiring us to develop a more efficient, selective information routing framework for VAD.

To this end, we propose \textbf{ReCoVAD}, a dual-pathway framework that enhances model collaboration to construct the loop of efficient “\textbf{Re}flex” filtering with deeper “\textbf{Co}nscious” reasoning in Figure \ref{fig:teaser}(b). The reflex pathway leverages a lightweight CLIP \cite{radford2021learning} model to fuse visual features with prompts derived from textual event prototypes, producing a decision vector to query a dynamic memory of representative frames' records (decision vectors- detection scores). If a frame falls within the coverage of the memory, it is regarded as redundant, and its anomaly score is retrieved directly, emulating low-latency, reflex filtering. Frames that fail this check are escalated to the conscious pathway, where a medium-scale (7B) VLM generates textual event descriptions and anomaly scores under the guidance of the textual event prototypes. These outputs are then written back into the reflex pathway's memory, progressively refining the system’s internal representations. To complete the feedback loop, the conscious pathway also includes an LLM-based reasoner that periodically revisits the generated descriptions to identify novel anomalies, revise earlier decisions, and adapt the prototype prompts, thus realizing a top-down refinement coupled with bottom-up filtering.

Therefore, our contributions are summarized as follows:
\begin{enumerate}
\item[(1)]To our best knowledge, we are the first to propose a biological-inspired VAD framework that simulates the reflex–conscious pathways of the human nervous system, significantly reducing the computational burden of using VLM/LLM while maintaining detection performance. 
\item[(2)]In this framework, we establish a novel closed-loop architecture that couples bottom-up filtering with top-down refinement: a lightweight reflex path rapidly filters redundant frames via memory querying, while a VLM-LLM conscious path processes novel frames and dynamically refines memory and prompts to enable a self-evolving VAD system.
\item[(3)]
Through our extensive experiments, we demonstrate that dense-frame inference is unnecessary for large-model-based VAD. Our ReCoVAD achieves state-of-the-art, training-free performance on the UCF-Crime and XD-Violence datasets, while using only \textbf{28.55\%} and \textbf{16.04\%} of the frames compared to previous large-model-based approaches.
\end{enumerate}

\section{Related Work}

\subsection{Video Anomaly Detection}

Video Anomaly Detection (VAD) methods are commonly grouped into one-class, unsupervised, and weakly supervised settings, based on the availability of anomalies and labels during training. One-class VAD assumes access to only normal videos and models normal patterns via direct statistical modeling \cite{benezeth2009abnormal,cheng2015video,hirschorn2023normalizing} or implicitly through proxy tasks \cite{hirschorn2023normalizing,liu2018future,xu2019video,shi2015convolutional,singh2018deep,liu2022learning,park2020learning,li2020spatial,tao2024feature,fang2020multi}, including future frame prediction \cite{liu2018future, xu2019video}, reconstruction \cite{shi2015convolutional, park2020learning}, and contrastive learning \cite{li2020spatial, tao2024feature}. Unsupervised VAD, on the other hand, assumes a mixture of normal and abnormal videos without labels \cite{thakare2023dyannet,thakare2023rareanom,tur2023unsupervised}, often relying on the assumption that normal events dominate and applying similar modeling of normal data in one-class methods. Methods, \textit{e.g.}GCL\cite{zaheer2022generative}, further enhance the normal/abnormal distinction with cooperative training that uses pseudo-labels from autoencoder reconstruction errors to guide discriminators. Despite their scalability, these methods are prone to high false positives, as rare but normal events are often misclassified as anomalies \cite{urdmu, umil}, especially in open-world settings with diverse scenes. Weakly supervised VAD utilizes video-level anomaly labels, typically via Multiple Instance Learning (MIL) \cite{sultani2018real}. Many approaches enhance MIL with feature discrimination \cite{tian2021weakly, chen2023mgfn} or self-training \cite{zhong2019graph, li2022self, shi2023abnormal, hu2025seva}, yet they remain limited by their dependence on training-time anomaly types, which often differ from those encountered at test time. In summary, traditional methods rely heavily on distributional assumptions or supervision, making them brittle when faced with semantically diverse or unseen anomalies.

\subsection{LVLM/LLM in Video Anomaly Detection}

To address these limitations, recent works integrate large-scale pre-trained LVLMs and LLMs into the VAD \cite{zanella2024harnessing,yang2024follow,kim2023unsupervised,tang2024hawk,du2024uncovering}. These models possess broad prior knowledge and human-like reasoning capability acquired from large-scale pretraining and demonstrate strong transferability. LAVAD \cite{zanella2024harnessing} developed a training-free VAD that requires no training videos or annotations. The method uses the VLM to produce textual descriptions for densely sampled video frames, and then LLMs are prompted to play the role of law enforcement to analyze and produce anomaly degrees about these descriptions. The AnomalyRuler \cite{yang2024follow} further involves few-shot training videos to induce rules for the LLM to deduce anomalies during testing. With the strong capability, recent works have also developed new tasks in VAD using LVLM/LLM, such as abnormal event Q\&A in HAWK \cite{tang2024hawk} and anomaly cause analysis in CUVA \cite{du2024uncovering}. However, these new paradigms with large pretrained models also introduce challenges in high inference cost and efficiency gaps that our proposed method seeks to address.

\section{Methodology}

\subsection{Problem Setup}
In this paper, we study the challenging task of training-free video anomaly detection (VAD). Given a test video, the objective is to assign an anomaly likelihood score $s \in [0,1]$ to each frame, indicating the probability of abnormal behavior. Unlike conventional approaches that rely on learning from the training set, training-free VAD \cite{zanella2024harnessing} requires detecting anomalies in test videos without any access to training data. Formally, let a test video be represented as a sequence of $T$ frames $V = \{I_1, I_2, ..., I_T\}$ sampled at a fixed rate. To perform inference, recent methods leverage a large vision-language model (VLM) $\Phi_{\text{VLM}}$ and a large language model (LLM) $\Phi_{\text{LLM}}$ to compute the anomaly scores $p$ according to the following standard pipeline:
\begin{equation}
\begin{aligned}
        C_I=\Phi_{VLM}(I)\space,
    s=\Phi_{LLM}(C_I), I \in V,
\end{aligned}
\label{eq:lavad}
\end{equation}
where $\Phi_{VLM}$ generates the caption $C_I$ to describe the events in the frame $I$, and $\Phi_{LLM}$ reasons on $C_I$ to determine whether these events are anomalies. This approach of invoking large models on dense video frame sequences in high computational costs, limiting VLM/LLM-based methods in the real-world VAD application.

In this paper, we will show that this hard labor of large pretrained models is unnecessary. To demonstrate this, we construct a dual-pathway architecture \textbf{ReCoVAD}, simulating how the human nervous system allocates cognitive resources selectively. This architecture allows the $\Phi_{VLM}$ and $\Phi_{LLM}$ to operate only on a small but core subset of the original frame sequence, while still achieving the state-of-the-art performance.

\subsection{Initialization}
\begin{figure}[t]
    \centering
     \includegraphics[width=1.0\linewidth]
     {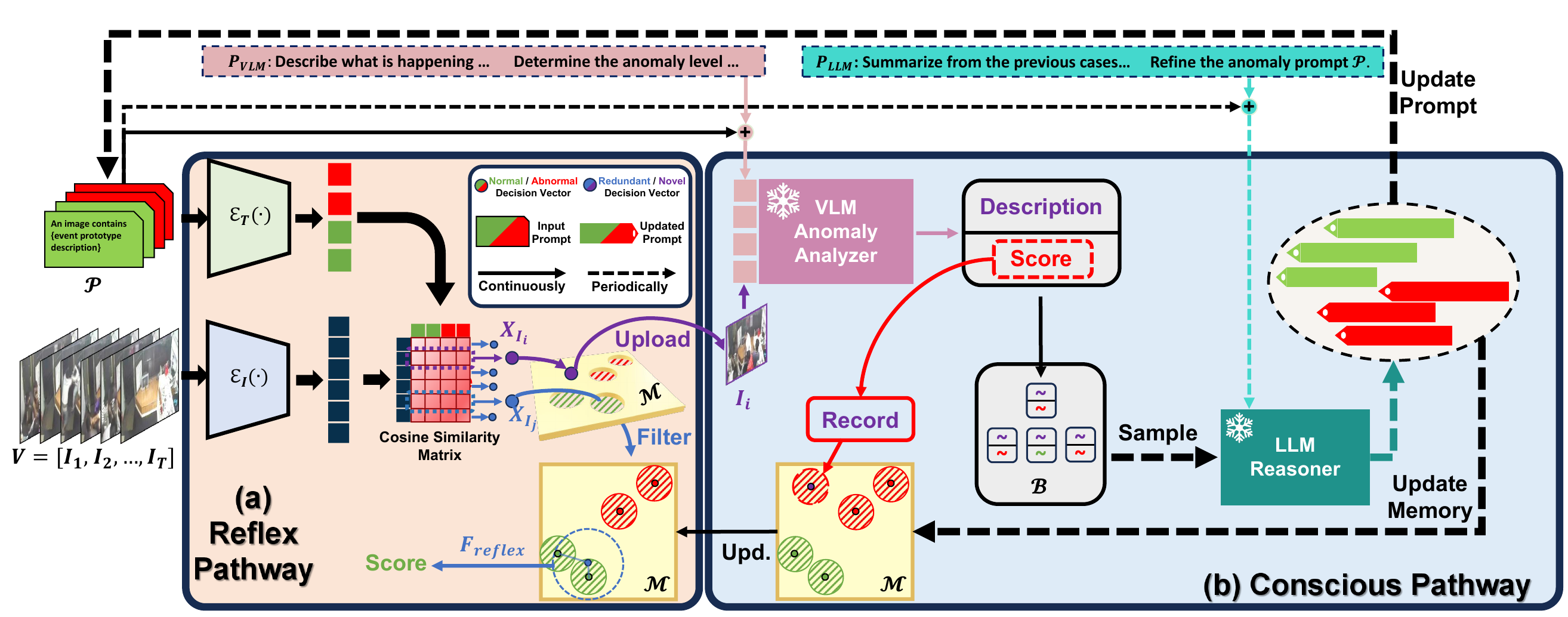}
     \caption{\textbf{ReCoVAD} consists of a \textbf{Reflex} and a \textbf{Conscious} pathway. The \textbf{Reflex} pathway employs a lightweight CLIP model to construct decision vectors $X_I$ by matching frame $I$’s visual features with textual event prototypes in the knowledge prompt $\mathcal{P}$. It then queries a dynamic memory $\mathcal{M}$ of representative records to decide if deep analysis is needed. If not, the anomaly score is retrieved directly via the reflex function based on the frame's neighbors in $\mathcal{M}$. Otherwise, the \textbf{Conscious} pathway processes the frame using a VLM anomaly analyzer to generate event descriptions and anomaly scores under the guidance of $\mathcal{P}$, updating $\mathcal{M}$ with new records and contributing to the description set $\mathcal{B}$. An LLM-based reasoner then periodically samples from $\mathcal{B}$ to revise prior decisions and refine $\mathcal{P}$, which in turn guides both pathways for top-down refinement.}

     \label{fig:pipeline}
\end{figure}

The overall framework of the ReCoVAD is illustrated in Figure \ref{fig:pipeline}. It adopts a multi-model architecture that comprises two complementary pathways: the \textbf{Reflex} pathway and \textbf{Conscious} pathway. The information transition between the two pathways is completed through the knowledge prompt $\mathcal{P}$, a list of $L$ protocol event descriptions formulated as \textit{person does something in some place} ($L$ is the total number of descriptions in $\mathcal{P}$). Although it is updated by the conscious pathway, manual initialization is required. The initial $\mathcal{P}$ is set to contain $3$ brief descriptions for both the normal and abnormal actions. The number of descriptions is updated into $L$ in the first round of LLM reasoning in Section \ref{sec:llm}. Due to the length limit, we put the full initiation of $\mathcal{P}$ in the Appendix. 

\subsection{Reflex Pathway}

The reflex pathway draws inspiration from the human's reflex arc to filter out frames that can be addressed with the previous records with function $F_{filter}$, thereby enabling direct anomaly detection using the function $F_{reflex}$ while uploading frames that require deep processing. 

To build the foundation for $F_{filter}$ and $F_{reflex}$, we simulate how human reflexes are shaped by both innate perception and summarized knowledge, constructing a decision space by combining input frame features with the knowledge prompt $\mathcal{P}$ (derived from initialization and LLM reasoning) in the aligned vision-language feature space of a light-weight CLIP model. Specifically, for the frame $I$ in the test frame sequence $V$, we use the visual encoder $\mathcal{E}_I(\cdot)$ of CLIP to extract its visual representation $\mathcal{E}_I(I)$. Meanwhile, we also formulate $\mathcal{P}$ as a set of normal/abnormal event prototype descriptions:$\{ p_i \}_{i=1}^{L}$. After processing through CLIP's text encoder $\mathcal{E}_T(\cdot)$, we obtain the corresponding text representations: $\{ \mathcal{E}_T(p_i) \}_{i=1}^{L}$. The decision vector $X_I \in \mathbb{R}^L$ is then produced through the cosine similarity between $\mathcal{E}_I(I)$ and $\{ \mathcal{E}_T(p_i) \}_{i=1}^{L}$:
\begin{equation}
    X_I=[\gamma\left \langle \mathcal{E}_I(I),\mathcal{E}_T(p_1) \right \rangle,\gamma\left \langle \mathcal{E}_I(I),\mathcal{E}_T(p_2) \right \rangle,...,\gamma\left \langle \mathcal{E}_I(I),\mathcal{E}_T(p_L) \right\rangle]^T, \label{eq:x}
\end{equation}
where $\left \langle\cdot,\cdot\right \rangle$ represents this cosine similarity and $\gamma$ is CLIP model's log scale factor. The decision vector helps reduce task-irrelevant information. Since pre-trained visual encoders often lack task-specific priors and extract unrelated elements, computing similarities with event prototypes allows us to localize the fame within the task space defined by $\mathcal{P}$, effectively denoise distractions.

After obtaining the decision vector $X_I$, we can describe the two core functions as follows:

\textbf{Filter Function $F_{filter}$}: The function $F_{filter}$ identifies whether the input frame $I$ requires deep analysis by determining if it falls within the coverage of a dynamic memory $\mathcal{M}$ storing the representative prediction records. Each record is formulated as a dictionary-like data frame to describe an input-detection pair:
\begin{equation}
    r=\{\textit{"visual"}:\mathcal{E}_I(I),\textit{"decision vector"}:X_I,\textit{"s"}:s\},
\end{equation} 
where $s$ is the recorded anomaly score of the frame $I$ processed from the raw output from the conscious path in Section \ref{sec:llm}.

To fulfill the frame filtering, the memory $\mathcal{M}$ should be able to (1) effectively cover the main distribution of the frame stream with only a limited amount of frame records to reduce the request for large model inference; (2) establish a metric to determine whether the input frame fits in its coverage.
To construct such $\mathcal{M}$, we adapt the idea from the greedy sampling from the core set algorithm \cite{guo2022deepcore,yang2023towards,roth2022towards}. For each new frame $I$ with its $X_I$, if $\mathcal{M}$ is empty, we add its corresponding record to $\mathcal{M}$ directly. Otherwise, we use $F_{filter}$ to decide the next move:
\begin{equation}
    F_{filter}(I,\mathcal{M})=\mathbb{I}_{\{\min_{r \in \mathcal{M}}D(X_I,r[\textit{"decision vector"}])>\epsilon  \}},
\end{equation}
$F_{filter}$ calculate $X_I$'s shortest distance to existing decision vectors in the memory $\mathcal{M}$ with metric $D(\cdot,\cdot)$ (\textit{e.g.} Euclidean distance). If the distance exceeds a predefined threshold $\epsilon$, $F_{filter}$ flags $1$ and the frame is considered as a novel sample. It is sent to the large pretrained models in the conscious pathway for further analysis. The resulting anomaly score $s$ from the conscious pathway is averaged with $K$-nearest neighbors to mitigate the noise in the large model’s predictions. Finally, the score $s$, the decision vector $X_I$, and the visual feature of the frame $I$ are assembled as a new record $r$, added into memory $\mathcal{M}$. If $X_I$ falls within the coverage of the memory $\mathcal{M}$, $F_{filter}$ flags $0$. The corresponding frame $I$ is directly processed by the reflex function $F_{reflex}$. 

Through $F_{filter}$, each record in $\mathcal{M}$ defines a hypersphere of radius $\epsilon$ around its decision vector, covering a local region. This ensures that all previously seen frames are either selected into $\mathcal{M}$ or covered by at least one hypersphere. Thus, $\mathcal{M}$ achieves compact coverage of known data while enabling novelty detection based on coverage failure. The radius $\epsilon$ controls the coverage of each record in the memory and thereby adjusts the proportion of frames routed to the conscious pathway.

\textbf{Reflex Function $F_{reflex}$}:
The reflex function $F_{reflex}$ computes the anomaly score of a video frame $I$ based on its decision vector $X_I$ and the memory $\mathcal{M}$. Since $F_{filter}$ has determined that the frame fits in the coverage of $\mathcal{M}$, its anomaly score is inferred from its neighbors in $\mathcal{M}$. In practice, we find it effective to consider neighbors within the decision hypersphere of the radius $a\cdot \epsilon$ around decision vector $X_I$. Also, we observe that large models are sensitive to signals of abnormal events, leading to potential false positives. To improve robustness, we make the anomaly decision more conservative: a frame is labeled as anomalous only if all neighbors within the $a\cdot\epsilon$ radius range are also labeled as anomalous. Accordingly, the final anomaly score is defined as the minimum score among neighbors of the frame in the memory $\mathcal{M}$:
\begin{gather}
F_{reflex}(I, \mathcal{M}) = \min_{r_j \in \mathcal{N}(I)} (r_j[\textit{"s"}]),\\
\mathcal{N}(I) = \{ r_j \in \mathcal{M} \mid \|X_I - r_j[\textit{"decision vector"}]\|_2 < a\cdot\epsilon \},
\end{gather}
where $\mathcal{N}(I)$ denotes the neighbor records of $I$ that falls within the decision hypersphere, and $r_j[\textit{"s"}]$ is the recorded anomaly score in $r_j$. The score from $F_{reflex}$ is then finalized as the output with a window smoothing, where the score is averaged with scores of the $C$ temporal nearest frames to improve the prediction consistency.
 
\subsection{Conscious Pathway}
\label{sec:llm}

The conscious pathway leverages large models to process novel video frames uploaded by the reflex pathway, mimicking complex human cognitive processes. It consists of two modules: (1) a VLM anomaly analyzer to describe the events and determine the anomaly score of the uploaded frames; (2) an LLM model that summarizes from the previous cases by the VLM anomaly analyzer to refine the knowledge prompt $\mathcal{P}$.

\textbf{VLM Anomaly Analyzer}: In this module, leveraging the knowledge prompt $\mathcal{P}$ and structured instructions, we integrate event description and anomaly detection into one midden-scale VLM (7B) $\Phi_{VLM}$, reducing the computational burden and information loss of the two-stage pipeline in Eq. \ref{eq:lavad}.

Given a video frame $I$ and the knowledge prompt $\mathcal{P}$, we construct the instruction prompt $P_{VLM}$ to guide the VLM anomaly analyzer to: (1) describe events in the frame, (2) compare them with prototypes in $\mathcal{P}$ and (3) assign an anomaly score chosen from the option list: \textit{OPTIONS}. The \textit{OPTIONS} contains $9$ distinct anomaly scores (real numbers from 0 to 1). Each score in \textit{OPTIONS} is compiled with its explanation, which is based on a match/mismatch between the event and the normal/abnormal prototypes in $\mathcal{P}$, thereby reducing the arbitrary decisions of VLM $\Phi_{VLM}$. Finally, $P_{VLM}$ specifies output format as $(des, s)$  where $des$ is the detailed description of the events in the frame and $s$ is the anomaly score used to support the update of the new record for memory $\mathcal{M}$. The $(des,s)$ pair is then appended to a temporal description set $\mathcal{B}$, used to support the LLM to summarize and refine the knowledge prompt $\mathcal{P}$. Therefore, the prompt for the VLM $\Phi_{VLM}$ is the concatenation of $,P_{VLM}$ and $\mathcal{P}$: $P_{VLM}\circ \mathcal{P}$. The details of $P_{VLM}$ and \textit{OPTIONS} can be found in the Appendix. The output in this pathway is as follows:

\textbf{LLM Reasoner:} Finally, the LLM analyzes accumulated $(des, s)$ pairs to emulate human-like conscious reasoning, enabling refinement of $\mathcal{P}$ and thereby enhancing both $\Phi_{VLM}$'s anomaly detection and the construction of decision vectors in the reflex pathway. Specifically, we randomly sample a subset $\mathcal{B}^{'}\subseteq\mathcal{B}$ of $b$ pairs, and design the prompt $P_{LLM}$ asking the LLM to clarify the previous $\mathcal{P}$ to describe the normal/abnormal events better. The details of the $P_{LLM}$ can be found in the Appendix.
Therefore, the  $\mathcal{P}$ can be updated with the concatenation of $P_{LLM}$, $\mathcal{P}$ and $\mathcal{B}^{'}$:
\begin{equation}
    \mathcal{P}\leftarrow\Phi_{LLM}(P_{LLM}\circ \mathcal{P}\circ \mathcal{B}^{'}).
\end{equation}
The LLM is only called at a fixed interval of $N$ videos to accumulate sufficient cases for analysis. The updated $\mathcal{P}$ is then fed back to the reflex pathway and $\Phi_{VLM}$ for adjustment. 

\begin{algorithm}
\caption{ReCoVAD Framework}
\renewcommand{\algorithmicrequire}{\textbf{Input:}}
\renewcommand{\algorithmicensure}{\textbf{Output:}}
\begin{algorithmic}[1]
\REQUIRE Test Video set $\mathcal{V}$ , image/text encoder $\mathcal{E}_I(\cdot)$/$\mathcal{E}_T(\cdot)$, large pretrained models $\Phi_{VLM}$, $\Phi_{LLM}$, initial prompt $\mathcal{P}$.
      \ENSURE Anomalous scores $s$ for each frame.
      \STATE Initialize $\mathcal{M}=\emptyset$, $\mathcal{B}=\emptyset$, $n=0$;
      \FOR{each video $V$ in video set $\mathcal{V}$}
      \FOR{each frame $I$ in $V$}
        \STATE Formulate $\mathcal{P}$ as $\{p_i\}_{i=1}^L$ and calculate $X_I$ with Eq.~\ref{eq:x};
        \IF{$F_{filter}(I,\mathcal{M})==1$}
          \STATE $(\textit{des}, s)=\Phi_{VLM}(I, P_{VLM}\circ \mathcal{P})$;
          \STATE $r \leftarrow \{\textit{visual}: \mathcal{E}_I(I),\, \textit{decision vector}: X_I,\, \textit{s}: s\}$;
          \STATE Add $r$ to $\mathcal{M}$; add $(\textit{des}, s)$ to $\mathcal{B}$;
        \ELSE
          \STATE $s \leftarrow F_{reflex}(I,\mathcal{M})$;
        \ENDIF
        
      \ENDFOR
      \STATE $n \leftarrow n+1$;
        \IF{$n \bmod N == 0$}
          \STATE Sample $b$ records from $\mathcal{B}$ to form the subset $\mathcal{B}^{'}$;
          \STATE Update $\mathcal{P} \leftarrow \Phi_{LLM}(P_{LLM}\circ \mathcal{P}\circ \mathcal{B}^{'})$;
          \STATE Update $X_I$ for $\forall r \in \mathcal{M}$ with new $\mathcal{P}$ by Eq.~\ref{eq:x};
          \STATE Re-evaluate $s$ for the previous frame using $F_{reflex}$.
          \STATE $\mathcal{B} \leftarrow \emptyset$;
        \ENDIF
    \ENDFOR
\end{algorithmic}
\end{algorithm}
Notably, we update the memory $\mathcal{M}$ to fit the change, where we recalculate the decision vector of every $r\in \mathcal{M}$ with recorded visual features and  $\{p_i\}_{i=1}^L$ formulated from updated $\mathcal{P}$ using Eq. \ref{eq:x}. Also, we re-evaluate the anomaly scores with $F_{reflex}$ for the historic frames with the new memory $\mathcal{M}$ to correct previous mistakes, improving the overall accuracy. After the feedback, $\mathcal{B}$ is emptied for the next round to ensure the LLM always learns from the new events. 

In summary, the framework forms a top-down refinement coupled with bottom-up filtering, demonstrated as Algorithm 1.

\section{Experiments}

We evaluate our framework on UCF-Crime \cite{sultani2018real} and XD-Violence \cite{wu2020not}, focusing on Training-free VAD accuracy and the number of frames processed by large pretrained models. Unlike prior methods relying on dense frame inference, our approach achieves higher performance with only a sparse subset of the frame used by the previous methods, demonstrating strong efficiency and effectiveness.

\begin{wraptable}{r}{0.65\textwidth}  
\centering

\caption{Comparison against one-class, unsupervised, and training-free methods on UCF-Crime dataset}
\resizebox{\linewidth}{!}{
\begin{tabular}{l|c|c|c}
\hline
Method        & Supervised mod & AUC(\%)        & Frames for VLM or LLM \\ \hline
BODS \cite{wang2019gods}          & one-class      & 68.26          & -                        \\
GODS \cite{wang2019gods}           & one-class      & 70.46          & -                        \\
GCL \cite{zaheer2022generative}           & unsupervised   & 74.20          & -                        \\
DYANNET \cite{zaheer2022generative}       & unsupervised   & 79.26          & -                        \\
ZS CLIP \cite{zanella2024harnessing}       & Training-free  & 53.16          & -                        \\
Qwen2.5-VL    & Training-free  & 79.22          & 69,344                    \\
LAVAD \cite{zanella2024harnessing}         & Training-free  & 80.28          & 69,344                    \\
\textbf{ReCoVAD} & Training-free  & \textbf{82.28} & \textbf{19,797}          \\ \hline
\end{tabular}}
\label{tab:ucf}

\centering
\caption{Comparison against one-class, unsupervised, and training-free methods on XD-Violence dataset}
\resizebox{\linewidth}{!}{
\begin{tabular}{l|c|c|c|c}
\hline
Method        & Supervised mod & AP(\%)         & AUC(\%)        & Frames for VLM or LLM \\ \hline
HASAN et al. \cite{hasan2016learning}  & one-class      & -              & 50.32          &                          \\
LU et al. \cite{lu2013abnormal}     & one-class      & -              & 53.56          &                          \\
BODS \cite{wang2019gods}           & one-class      & -              & 57.32          & -                        \\
GODS \cite{wang2019gods}           & one-class      & -              & 61.56          & -                        \\
RAREANOM \cite{thakare2023rareanom}     &Unsupervised      & -        &68.33          & -                        \\
ZS CLIP \cite{zanella2024harnessing}      & Training-free  & 17.83          & 38.21          & -                        \\
Qwen2.5-VL    & Training-free  & 55.21          & 83.59          & 145,649                   \\
LAVAD \cite{zanella2024harnessing}         & Training-free  & 62.01          & 85.36          & 145,649                   \\
\textbf{ReCoVAD} & Training-free  & \textbf{65.66} & \textbf{86.38} & \textbf{23,362}           \\ \hline
\end{tabular}}
\label{tab:xd}
\end{wraptable}

\subsection{Experimental Setups}
\label{sec:exp}
\textbf{Dataset and Test Settings:} We evaluate our framework on the UCF-Crime \cite{sultani2018real} and XD-Violence \cite{wu2020not} datasets, which consist of real-world surveillance footage with diverse events. \textbf{UCF-Crime} includes 1,900 untrimmed videos covering 13 abnormal events. The training set has 800 normal and 810 abnormal videos, while the test set contains 140 normal and 150 abnormal videos with frame-level annotations. \textbf{XD-Violence} contains 4,754 multi-modal videos from movies and YouTube, spanning 6 types of violent events, with 3,954 training and 800 test videos. We follow LAVAD \cite{zanella2024harnessing}’s experimental setup, testing on the test sets without using training data or annotations. We follow the experimental setup of LAVAD \cite{zanella2024harnessing}, evaluating our framework on the test sets of both datasets without seeing or training with training videos or annotations. Notably, we randomly shuffle the test sets, which originally contain consecutive samples of the same event type. This shuffling increases task difficulty by preventing the model from exploiting event continuity, providing a more rigorous assessment of our framework’s ability to generalize to unseen scenes and events, and better simulating real-world conditions.

\textbf{Performance Metrics: }Following standard practice, we report the frame-level ROC AUC on UCF-Crime and XD-Violence datasets, which is regarded as a fair metric for class-imbalanced tasks like anomaly detection \cite{sultani2018real,liu2018future}. Also, we evaluate the frame-level average precision (AP), i.e., the area under the precision-recall curve, on XD-Violence following the setting in the work \cite{wu2020not,urdmu}

\textbf{Implementation Details:} The basic video inputs are represented as frame sequences sampled from each video every $16$ frames following LAVAD \cite{zanella2024harnessing,wu2024open}. The CLIP model in the reflex pathway is set as pretrained CLIP-ViT-B/16 \cite{radford2021learning}. The parameter $\epsilon$ in the memory $\mathcal{M}$ is set as $2.0$ for UCF-Crime and $4.0$ for XD-Violence to ensure the total compressed rate of the inputs falls in the reasonable range of $15\%-30\%$. The parameters for neighbor number $K$, the window size $C$, and the radius coefficient $a$ in $F_{reflex}$ are set uniformly for both datasets as $16$,$4$, and $2$. In the conscious pathway, we implement the VLM event describer as the Qwen2.5-VL-7B \cite{bai2025qwen25vltechnicalreport} model that is widely used by today's open-source community. At last, we set LLM as the Deepseek-V3 \cite{liu2024deepseek} model. The length $L$ for $\mathcal{P}$ that is summarized by the LLM is set to be $20$, with a half-to-half split for normal and abnormal events. We also set $N$ as 10, meaning the framework performs a summarization and
refinement process after observing every 10 videos. The parameter $b$ for the size of the subset $\mathcal{B}^{'}$ is set as 90. In practice, we sample $b/2$ $(des,s)$ pairs with score $s>0.5$ and $b/2$ pairs with $s<0.5$ to ensure normal/abnormal balance in $\mathcal{B}^{'}$. In the initialization round, when the total length of $\mathcal{P}$ is set as $3$, we shrink $\epsilon$ as $1.2$ uniformly at the beginning to ensure sufficient sampling for the LLM to analyze. After the first round of the LLM reasoning, $\epsilon$ is set to be the predefined value in each dataset.

\subsection{Experimental Analysis and Visualization}
\label{sec:results}

\begin{figure}[t]
    \centering
        \setlength{\belowcaptionskip}{-0.4cm} 
\setlength{\abovecaptionskip}{-0.1pt} 
     \includegraphics[width=1.0\linewidth]
     {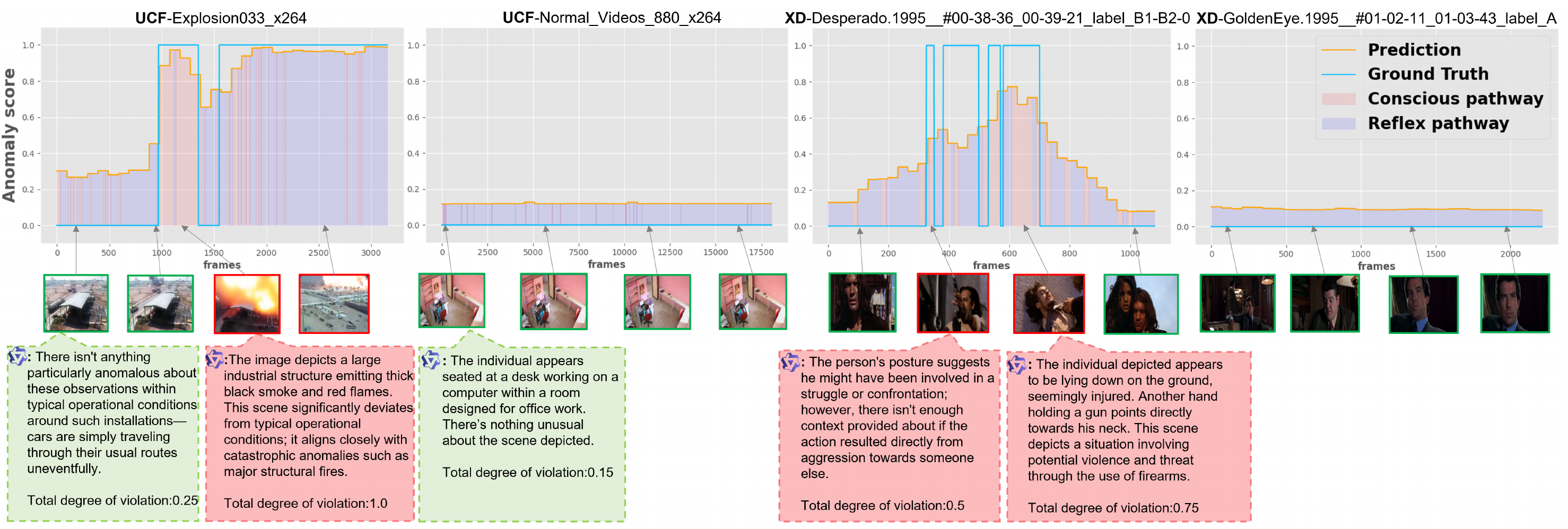}
     \caption{Visualization of the predictions made by the reflex pathway and the conscious pathway.}
     \label{fig:v_p}
\end{figure}
\textbf{Experimental Analysis: } The experimental results are presented in Table \ref{tab:ucf} and Table \ref{tab:xd}. In the tables, one-class methods are trained using only the normal videos from the training set, while unsupervised methods use the entire training set without access to any annotations. Training-free methods do not utilize any training data and perform inference directly on the test set. Notably, the “ZS CLIP”  refers to a zero-shot CLIP-based approach that computes anomaly scores by contrasting prompts for normal and anomalous events, following the same prompt design used in LAVAD \cite{zanella2024harnessing}.

As shown in Table \ref{tab:ucf} and Table \ref{tab:xd}, our method significantly outperforms traditional one-class and unsupervised video anomaly detection approaches. Most importantly, our framework achieves notable improvements in both accuracy and efficiency compared to previous training-free methods based on large pretrained models. Specifically, compared with the state-of-the-art training-free method LAVAD \cite{zanella2024harnessing}, our framework achieves a 2.00\% AUC improvement on UCF-Crime, a 3.65\% AP gain, and a 1.02\% AUC gain on XD-Violence. In addition to the superior performance, our method significantly reduces the computational burden on large pretrained models. Our approach processes only 28.55\% and 16.04\% of the frames used by LAVAD in UCF-Crime and XD-Violence with large pre-trained models. To further validate the effectiveness of the reflex pathway and the top-down refinement of the conscious pathway, we design a baseline using Qwen2.5-VL that predicts all video frames using the VLM anomaly analyzer $\Phi_{VLM}$ with $P_{VLM}$, without the reflex pathway or the top-down refinement. Compared to this baseline, our method achieves a 3.06\% AUC improvement on UCF-Crime, and on XD-Violence, gains of 10.45\% in AP and 2.79\% in AUC, while achieving computational savings on the large pretrained models. These results confirm that our framework delivers both superior detection performance and low computational burden for the large pretrained models.

\textbf{Visualization}: We further validate the roles of both pathways through prediction score visualizations. As demonstrated in Figure \ref{fig:v_p}, the reflex pathway (light purple regions) dominates the output scores, especially for routine/repetitive events in normal videos, effectively reducing reliance on the large model. For novel events like the explosion in \texttt{UCF-Explosion033\_x264} case, the reflex pathway identifies unfamiliar patterns and activates the conscious pathway for accurate description and anomaly detection. Once the anomaly is recorded, similar events in future frames are handled directly by the reflex pathway, minimizing repeated inference. This illustrates the coordination between pathways: the reflex pathway offers responses for the most redundant inputs, while the conscious pathway only handles novel cases, greatly reducing the computational burden of the large models.

\section{Ablation Study}

\subsection{Effectiveness of Components}
We evaluate the effectiveness of pathways in our framework through controlled ablations. Specifically, we isolate the reflex pathway and the conscious pathway, keeping one fixed while testing the other.
\begin{table*}
\setlength{\abovecaptionskip}{-1pt}
\begin{minipage}[t]{0.48\textwidth}
\caption{The effectiveness of the conscious pathway's components}
\resizebox{\linewidth}{!}{
\begin{tabular}{ccc|c}
\hline
Feedback $\mathcal{P}$ to reflex & Feedback $\mathcal{P}$ to $\Phi_{VLM}$ & \textit{OPTIONS} & AUC(\%) \\ \hline
\ding{55}                          & \ding{55}                           & \ding{55}                      & 70.80   \\
\ding{55}                          & \ding{51}                           & \ding{51}                      & 74.62   \\
\ding{51}                          & \ding{55}                           & \ding{55}                      & 77.83   \\
\ding{51}                          & \ding{51}                           & \ding{55}                      & 79.92   \\
\ding{51}                          & \ding{51}                           & \ding{51}                      & 82.28   \\ \hline
\end{tabular}}
\label{tab:ab_c}
\end{minipage}
\hfill
\begin{minipage}[t]{0.48\textwidth}
\caption{The effectiveness of the reflex pathway's components}
\resizebox{\linewidth}{!}{
\begin{tabular}{ccc|c}
\hline
$F_{reflex}$ & Use minimum among neighbors & Window smooth & AUC(\%) \\ \hline
\ding{55}                          & \ding{55}                           & \ding{55}                      & 79.24   \\
\ding{55}                          & \ding{55}                           & \ding{51}                      & 80.36   \\
\ding{51}                          & \ding{55}                           & \ding{55}                      & 79.97   \\
\ding{51}                          & \ding{51}                           & \ding{55}                      & 80.93   \\
\ding{51}                          & \ding{51}                           & \ding{51}                      & 82.28   \\ \hline
\end{tabular}}
\label{tab:ab_re}
\end{minipage}
\end{table*}

Table \ref{tab:ab_c} presents results for the conscious pathway. We examine whether the knowledge prompt $\mathcal{P}$, dynamically updated by the LLM $\Phi_{LLM}$, improves overall performance, and assess the role of both the knowledge prompt and the anomaly-level option list \textit{OPTIONS} in enhancing VLM $\Phi_{VLM}$'s detection. When $\mathcal{P}$ is blocked from both the reflex pathway and $\Phi_{VLM}$, performance drops significantly below the baseline. This is because $F_{filter}$ cannot construct decision vectors aligned with the current task space, leading to poor distance estimation in $F_{reflex}$ and ineffective frame selection. Additionally, when the $\Phi_{VLM}$ lacks updated knowledge, it also reduces detection accuracy. As shown in rows 2 and 3 of Table \ref{tab:ab_c}, enabling knowledge prompt $\mathcal{P}$ transmission to the VLM anomaly analyzer and reflex pathway improves performance by 7.03\% and 3.82\%, respectively. Rows 4 and 5 further evaluate the impact of \textit{OPTIONS}. In Row 4, we set the VLM to produce a raw anomaly score in $[0, 1]$ as in LAVAD when \textit{OPTIONS} is absent. In row 5, by incorporating \textit{OPTIONS}, the framework yields a 1.35\% gain, highlighting the benefit of providing interpretable anomaly cues.

We further examine the reflex pathway, focusing on $F_{reflex}$, as the functionality of $F_{filter}$ is validated in Section \ref{sec:results}. Table \ref{tab:ab_re} compares 3 testings: (1) Without $F_{reflex}$, the model uses the prediction of the nearest neighbor in memory $\mathcal{M}$ as output, resulting in a 3.04\% performance drop; (2) Rows 3 and 4 compare aggregation strategies in $F_{reflex}$: using the average vs.\ the minimum prediction score. The latter yields a 0.96\% improvement, better reducing false positives; (3) Applying a temporal smoothing window improves temporal consistency, boosting accuracy by 1.35\%  and 1.12\% in the framework with/without $F_{reflex}$, suggesting greater effectiveness when combined with $F_{reflex}$.

\subsection{Sensitivity to Hyper-parameters}

\begin{figure}[t]
\centering
\begin{minipage}{0.48\linewidth}
\centering
\resizebox{\linewidth}{!}{
\begin{tabular}{c|cccccc}
\hline
$\epsilon$ & 1.6    & 1.8    & 2.0    & 2.2    & 2.4    & 2.6    \\ \hline
AUC(\%)   & 77.09  & 80.01  & 82.28  & 81.45  & 77.59  & 77.97  \\
frames for $\Phi_{VLM}$ & 30,309 & 24,337 & 19,797 & 18,246 & 13,338 & 10,725 \\ \hline
\end{tabular}}
\captionof{table}{Ablation on the parameter $\epsilon$ in $F_{filter}$}
\label{tab:ep}
\end{minipage}
\hfill
\begin{minipage}{0.49\linewidth}
\centering
\includegraphics[width=\linewidth]{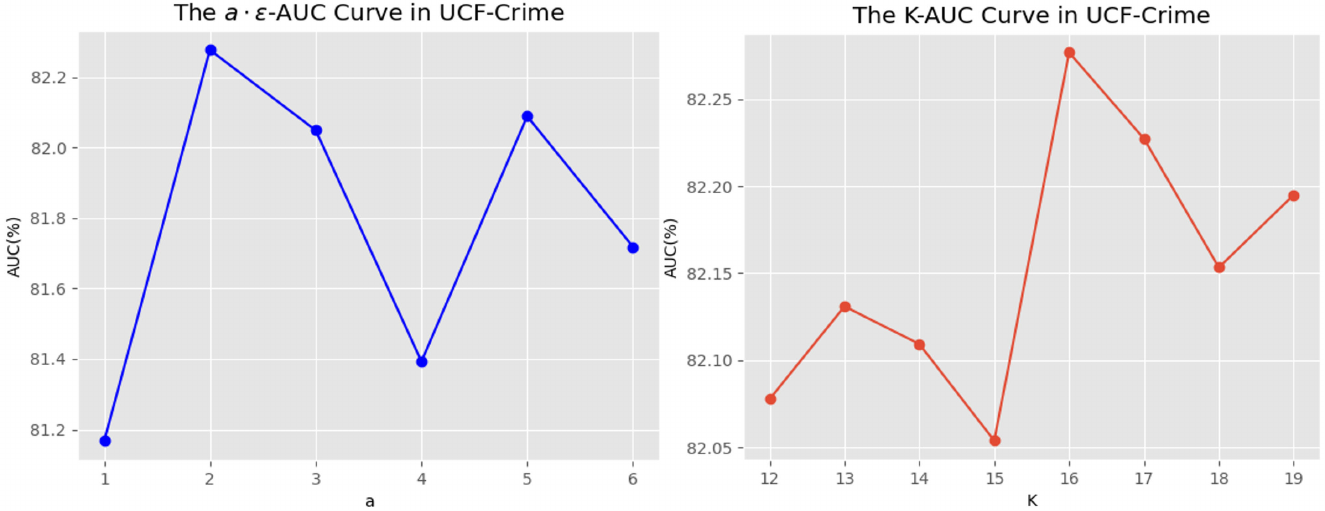}
\captionof{figure}{The ablation on parameter $K$ and the radius of the decision hypersphere in $F_{reflex}$}
\label{fig:abl}
\end{minipage}

\end{figure}

We further analyze the sensitivity of our framework to key hyperparameters. We first investigate $\epsilon$, which determines the volume of the memory $\mathcal{M}$. As shown in Table \ref{tab:ep}, the choice of $\epsilon$ is crucial. When $\epsilon$ is too large, new frames are rarely added to the memory $\mathcal{M}$, which reduces the total number of recorded frames. This leads to an inaccurate representation of the distribution of seen frames, making it difficult to filter input frames effectively and ultimately degrading performance. In contrast, a small $\epsilon$ leads to excessive storage of redundant frames, increasing noise sensitivity, and triggering unnecessary large-model inference.

We also examine two hyperparameters in $F_{reflex}$: the radius $a \cdot \epsilon$ of the decision hypersphere, and the number of neighbors $K$ used for score smoothing when appending new records. As illustrated in Figure \ref{fig:abl}, a small radius makes the model fragile to noisy entries, while a large one risks including irrelevant frames. Similarly, a small $K$ fails to suppress noise, whereas a large $K$ may bury the original score. Overall, performance remains stable (within 1.8\% and $0.23\%$), demonstrating strong robustness to these hyperparameters. More ablations on hyperparameters are included in the Appendix.

\section{Conclusion and Limitation}
\label{sec:con}

To summarize, in this paper, through our biological-inspired dual pathway ReCoVAD framework and extensive experiments, we demonstrate that the previous high-frequency reasoning with large pre-trained models in VAD is unnecessary. ReCoVAD reduces the computational burden of the VLM/LLM models by using a reflex pathway to filter and respond to redundant inputs, and leaves the large model to deal with novel inputs to refine the entire system. The experiments and visualization show that ReCoVAD achieves state-of-the-art performance, with the reflex pathway handling the majority of the inputs, greatly reducing the computational burden of the large models. However, our work still has the limitation that the current framework is built on a limited set of models. In future work, we plan to extend our study to a broader range of VLM/VLM choices to further investigate the generalization ability of our framework.
\section{Acknowledgment}
This work was supported by the Program of Beijing Municipal Science and Technology Commission Foundation
(No.Z241100003524010), in part by the National Natural
Science Foundation of China under Grant 62088102, in part
by AI Joint Lab of Future Urban Infrastructure, sponsored by
Fuzhou Chengtou New Infrastructure Group and Boyun Vision Co. Ltd, and in part by the PKU-NTU Joint Research
Institute (JRI) sponsored by a donation from the Ng Teng
Fong Charitable Foundation.

{
\bibliographystyle{abbrv}
\bibliography{main}

@String(ICME = {Int. Conf. Multimedia and Expo})

@String(AAAI = {AAAI})

@String(ICME  =	{ICME})

@article{liu2023generalized,
  title={Generalized video anomaly event detection: Systematic taxonomy and comparison of deep models},
  author={Liu, Yang and Yang, Dingkang and Wang, Yan and Liu, Jing and Liu, Jun and Boukerche, Azzedine and Sun, Peng and Song, Liang},
  journal={ACM Computing Surveys},
  year={2023},
  publisher={ACM New York, NY}
}

@inproceedings{benezeth2009abnormal,
  title={Abnormal events detection based on spatio-temporal co-occurences},
  author={Benezeth, Yannick and Jodoin, P-M and Saligrama, Venkatesh and Rosenberger, Christophe},
  booktitle={2009 IEEE conference on computer vision and pattern recognition},
  pages={2458--2465},
  year={2009},
  organization={IEEE}
}

@inproceedings{cheng2015video,
  title={Video anomaly detection and localization using hierarchical feature representation and Gaussian process regression},
  author={Cheng, Kai-Wen and Chen, Yie-Tarng and Fang, Wen-Hsien},
  booktitle={Proceedings of the IEEE Conference on Computer Vision and Pattern Recognition},
  pages={2909--2917},
  year={2015}
}

@inproceedings{hirschorn2023normalizing,
  title={Normalizing flows for human pose anomaly detection},
  author={Hirschorn, Or and Avidan, Shai},
  booktitle={Proceedings of the IEEE/CVF International Conference on Computer Vision},
  pages={13545--13554},
  year={2023}
}

@inproceedings{liu2018future,
  title={Future frame prediction for anomaly detection--a new baseline},
  author={Liu, Wen and Luo, Weixin and Lian, Dongze and Gao, Shenghua},
  booktitle={Proceedings of the IEEE conference on computer vision and pattern recognition},
  pages={6536--6545},
  year={2018}
}

@article{shi2015convolutional,
  title={Convolutional LSTM network: A machine learning approach for precipitation nowcasting},
  author={Shi, Xingjian and Chen, Zhourong and Wang, Hao and Yeung, Dit-Yan and Wong, Wai-Kin and Woo, Wang-chun},
  journal={Advances in neural information processing systems},
  volume={28},
  year={2015}
}

@inproceedings{hasan2016learning,
  title={Learning temporal regularity in video sequences},
  author={Hasan, Mahmudul and Choi, Jonghyun and Neumann, Jan and Roy-Chowdhury, Amit K and Davis, Larry S},
  booktitle={Proceedings of the IEEE conference on computer vision and pattern recognition},
  pages={733--742},
  year={2016}
}

@inproceedings{singh2018deep,
  title={A deep learning based technique for anomaly detection in surveillance videos},
  author={Singh, Prakhar and Pankajakshan, Vinod},
  booktitle={2018 Twenty Fourth National Conference on Communications (NCC)},
  pages={1--6},
  year={2018},
  organization={IEEE}
}

@inproceedings{liu2022learning,
  title={Learning appearance-motion normality for video anomaly detection},
  author={Liu, Yang and Liu, Jing and Zhao, Mengyang and Yang, Dingkang and Zhu, Xiaoguang and Song, Liang},
  booktitle={2022 IEEE International Conference on Multimedia and Expo (ICME)},
  pages={1--6},
  year={2022},
  organization={IEEE}
}

@inproceedings{park2020learning,
  title={Learning memory-guided normality for anomaly detection},
  author={Park, Hyunjong and Noh, Jongyoun and Ham, Bumsub},
  booktitle={Proceedings of the IEEE/CVF conference on computer vision and pattern recognition},
  pages={14372--14381},
  year={2020}
}

@inproceedings{openvad,
  title={Towards Open Set Video Anomaly Detection},
  author={Zhu, Yuansheng and Bao, Wentao and Yu, Qi},
  booktitle={European Conference on Computer Vision},
  pages={395--412},
  year={2022},
  organization={Springer}
}

@inproceedings{tian2021weakly,
  title={Weakly-supervised video anomaly detection with robust temporal feature magnitude learning},
  author={Tian, Yu and Pang, Guansong and Chen, Yuanhong and Singh, Rajvinder and Verjans, Johan W and Carneiro, Gustavo},
  booktitle={Proceedings of the IEEE/CVF international conference on computer vision},
  pages={4975--4986},
  year={2021}
}

@inproceedings{sultani2018real,
  title={Real-world anomaly detection in surveillance videos},
  author={Sultani, Waqas and Chen, Chen and Shah, Mubarak},
  booktitle={Proceedings of the IEEE conference on computer vision and pattern recognition},
  pages={6479--6488},
  year={2018}
}

@inproceedings{chen2023mgfn,
  title={Mgfn: Magnitude-contrastive glance-and-focus network for weakly-supervised video anomaly detection},
  author={Chen, Yingxian and Liu, Zhengzhe and Zhang, Baoheng and Fok, Wilton and Qi, Xiaojuan and Wu, Yik-Chung},
  booktitle={Proceedings of the AAAI Conference on Artificial Intelligence},
  volume={37},
  pages={387--395},
  year={2023}
}

@inproceedings{umil,
  title={Unbiased Multiple Instance Learning for Weakly Supervised Video Anomaly Detection},
  author={Lv, Hui and Yue, Zhongqi and Sun, Qianru and Luo, Bin and Cui, Zhen and Zhang, Hanwang},
  booktitle={Proceedings of the IEEE/CVF Conference on Computer Vision and Pattern Recognition},
  pages={8022--8031},
  year={2023}
}

@article{urdmu,
  title={Dual Memory Units with Uncertainty Regulation for Weakly Supervised Video Anomaly Detection},
  author={Zhou, Hang and Yu, Junqing and Yang, Wei},
  journal={arXiv preprint arXiv:2302.05160},
  year={2023}
}

@inproceedings{zhong2019graph,
  title={Graph convolutional label noise cleaner: Train a plug-and-play action classifier for anomaly detection},
  author={Zhong, Jia-Xing and Li, Nannan and Kong, Weijie and Liu, Shan and Li, Thomas H and Li, Ge},
  booktitle={Proceedings of the IEEE/CVF conference on computer vision and pattern recognition},
  pages={1237--1246},
  year={2019}
}

@inproceedings{lu2013abnormal,
  title={Abnormal event detection at 150 fps in matlab},
  author={Lu, Cewu and Shi, Jianping and Jia, Jiaya},
  booktitle={Proceedings of the IEEE international conference on computer vision},
  pages={2720--2727},
  year={2013}
}

@inproceedings{wang2019gods,
  title={Gods: Generalized one-class discriminative subspaces for anomaly detection},
  author={Wang, Jue and Cherian, Anoop},
  booktitle={Proceedings of the IEEE/CVF International Conference on Computer Vision},
  pages={8201--8211},
  year={2019}
}

@inproceedings{wu2020not,
  title={Not only look, but also listen: Learning multimodal violence detection under weak supervision},
  author={Wu, Peng and Liu, Jing and Shi, Yujia and Sun, Yujia and Shao, Fangtao and Wu, Zhaoyang and Yang, Zhiwei},
  booktitle={Computer Vision--ECCV 2020: 16th European Conference, Glasgow, UK, August 23--28, 2020, Proceedings, Part XXX 16},
  pages={322--339},
  year={2020},
  organization={Springer}
}

@article{hu2025beyond,
  title={Beyond Entropy: Region Confidence Proxy for Wild Test-Time Adaptation},
  author={Hu, Zixuan and Hu, Yichun and Li, Xiaotong and Tang, Shixiang and Duan, Ling-Yu},
  journal={arXiv preprint arXiv:2505.20704},
  year={2025}
}

@inproceedings{li2022self,
  title={Self-training multi-sequence learning with transformer for weakly supervised video anomaly detection},
  author={Li, Shuo and Liu, Fang and Jiao, Licheng},
  booktitle={Proceedings of the AAAI Conference on Artificial Intelligence},
  volume={36},
  pages={1395--1403},
  year={2022}
}

@inproceedings{hu2024lead,
  title={Lead: Exploring logit space evolution for model selection},
  author={Hu, Zixuan and Li, Xiaotong and Tang, Shixiang and Liu, Jun and Hu, Yichun and Duan, Ling-Yu},
  booktitle={Proceedings of the IEEE/CVF Conference on Computer Vision and Pattern Recognition},
  pages={28664--28673},
  year={2024}
}

@inproceedings{hu2025adaptive,
  title={Adaptive Dual Uncertainty Optimization: Boosting Monocular 3D Object Detection under Test-Time Shifts},
  author={Hu, Zixuan and Li, Dongxiao and Ma, Xinzhu and Tang, Shixiang and Li, Xiaotong and Yang, Wenhan and Duan, Ling-Yu},
  booktitle={Proceedings of the IEEE/CVF International Conference on Computer Vision},
  pages={7273--7283},
  year={2025}
}

@article{hu2025seva,
  title={SEVA: Leveraging Single-Step Ensemble of Vicinal Augmentations for Test-Time Adaptation},
  author={Hu, Zixuan and Hu, Yichun and Duan, Ling-Yu},
  journal={arXiv preprint arXiv:2505.04087},
  year={2025}
}

@inproceedings{wu2024open,
  title={Open-vocabulary video anomaly detection},
  author={Wu, Peng and Zhou, Xuerong and Pang, Guansong and Sun, Yujia and Liu, Jing and Wang, Peng and Zhang, Yanning},
  booktitle={Proceedings of the IEEE/CVF Conference on Computer Vision and Pattern Recognition},
  pages={18297--18307},
  year={2024}
}

@article{pang2021deep,
  title={Deep learning for anomaly detection: A review},
  author={Pang, Guansong and Shen, Chunhua and Cao, Longbing and Hengel, Anton Van Den},
  journal={ACM computing surveys (CSUR)},
  volume={54},
  number={2},
  pages={1--38},
  year={2021},
  publisher={ACM New York, NY, USA}
}

@inproceedings{roth2022towards,
  title={Towards total recall in industrial anomaly detection},
  author={Roth, Karsten and Pemula, Latha and Zepeda, Joaquin and Sch{\"o}lkopf, Bernhard and Brox, Thomas and Gehler, Peter},
  booktitle={Proceedings of the IEEE/CVF conference on computer vision and pattern recognition},
  pages={14318--14328},
  year={2022}
}

@inproceedings{radford2021learning,
  title={Learning transferable visual models from natural language supervision},
  author={Radford, Alec and Kim, Jong Wook and Hallacy, Chris and Ramesh, Aditya and Goh, Gabriel and Agarwal, Sandhini and Sastry, Girish and Askell, Amanda and Mishkin, Pamela and Clark, Jack and others},
  booktitle={International conference on machine learning},
  pages={8748--8763},
  year={2021},
  organization={PMLR}
}

@article{li2020spatial,
  title={Spatial-temporal cascade autoencoder for video anomaly detection in crowded scenes},
  author={Li, Nanjun and Chang, Faliang and Liu, Chunsheng},
  journal={IEEE Transactions on Multimedia},
  volume={23},
  pages={203--215},
  year={2020},
  publisher={IEEE}
}

@article{tao2024feature,
  title={Feature reconstruction with disruption for unsupervised video anomaly detection},
  author={Tao, Chenchen and Wang, Chong and Lin, Sunqi and Cai, Suhang and Li, Di and Qian, Jiangbo},
  journal={IEEE Transactions on Multimedia},
  year={2024},
  publisher={IEEE}
}

@article{xu2019video,
  title={Video anomaly detection and localization based on an adaptive intra-frame classification network},
  author={Xu, Ke and Sun, Tanfeng and Jiang, Xinghao},
  journal={IEEE Transactions on Multimedia},
  volume={22},
  number={2},
  pages={394--406},
  year={2019},
  publisher={IEEE}
}

@article{shi2023abnormal,
  title={Abnormal ratios guided multi-phase self-training for weakly-supervised video anomaly detection},
  author={Shi, Haoyue and Wang, Le and Zhou, Sanping and Hua, Gang and Tang, Wei},
  journal={IEEE Transactions on Multimedia},
  volume={26},
  pages={5575--5587},
  year={2023},
  publisher={IEEE}
}

@article{fang2020multi,
  title={Multi-encoder towards effective anomaly detection in videos},
  author={Fang, Zhiwen and Zhou, Joey Tianyi and Xiao, Yang and Li, Yanan and Yang, Feng},
  journal={IEEE Transactions on Multimedia},
  volume={23},
  pages={4106--4116},
  year={2020},
  publisher={IEEE}
}

@inproceedings{zanella2024harnessing,
  title={Harnessing large language models for training-free video anomaly detection},
  author={Zanella, Luca and Menapace, Willi and Mancini, Massimiliano and Wang, Yiming and Ricci, Elisa},
  booktitle={Proceedings of the IEEE/CVF Conference on Computer Vision and Pattern Recognition},
  pages={18527--18536},
  year={2024}
}

@inproceedings{yang2024follow,
  title={Follow the rules: reasoning for video anomaly detection with large language models},
  author={Yang, Yuchen and Lee, Kwonjoon and Dariush, Behzad and Cao, Yinzhi and Lo, Shao-Yuan},
  booktitle={European Conference on Computer Vision},
  pages={304--322},
  year={2024},
  organization={Springer}
}

@article{yao2020and,
  title={When, where, and what? A new dataset for anomaly detection in driving videos},
  author={Yao, Yu and Wang, Xizi and Xu, Mingze and Pu, Zelin and Atkins, Ella and Crandall, David},
  journal={arXiv preprint arXiv:2004.03044},
  year={2020}
}

@article{kim2023unsupervised,
  title={Unsupervised video anomaly detection based on similarity with predefined text descriptions},
  author={Kim, Jaehyun and Yoon, Seongwook and Choi, Taehyeon and Sull, Sanghoon},
  journal={Sensors},
  volume={23},
  number={14},
  pages={6256},
  year={2023},
  publisher={MDPI}
}

@article{tang2024hawk,
  title={Hawk: Learning to understand open-world video anomalies},
  author={Tang, Jiaqi and Lu, Hao and Wu, Ruizheng and Xu, Xiaogang and Ma, Ke and Fang, Cheng and Guo, Bin and Lu, Jiangbo and Chen, Qifeng and Chen, Yingcong},
  journal={Advances in Neural Information Processing Systems},
  volume={37},
  pages={139751--139785},
  year={2024}
}

@inproceedings{du2024uncovering,
  title={Uncovering what why and how: A comprehensive benchmark for causation understanding of video anomaly},
  author={Du, Hang and Zhang, Sicheng and Xie, Binzhu and Nan, Guoshun and Zhang, Jiayang and Xu, Junrui and Liu, Hangyu and Leng, Sicong and Liu, Jiangming and Fan, Hehe and others},
  booktitle={Proceedings of the IEEE/CVF Conference on Computer Vision and Pattern Recognition},
  pages={18793--18803},
  year={2024}
}

@inproceedings{samsi2023words,
  title={From words to watts: Benchmarking the energy costs of large language model inference},
  author={Samsi, Siddharth and Zhao, Dan and McDonald, Joseph and Li, Baolin and Michaleas, Adam and Jones, Michael and Bergeron, William and Kepner, Jeremy and Tiwari, Devesh and Gadepally, Vijay},
  booktitle={2023 IEEE High Performance Extreme Computing Conference (HPEC)},
  pages={1--9},
  year={2023},
  organization={IEEE}
}

@article{zhou2024survey,
  title={A survey on efficient inference for large language models, 2024},
  author={Zhou, Zixuan and Ning, Xuefei and Hong, Ke and Fu, Tianyu and Xu, Jiaming and Li, Shiyao and Lou, Yuming and Wang, Luning and Yuan, Zhihang and Li, Xiuhong and others},
  journal={URL https://arxiv. org/abs/2404.14294},
  year={2024}
}

@article{Meunier_2009,
   title={Hierarchical modularity in human brain functional networks},
   volume={3},
   ISSN={1662-5196},
   url={http://dx.doi.org/10.3389/neuro.11.037.2009},
   DOI={10.3389/neuro.11.037.2009},
   journal={Frontiers in Neuroinformatics},
   publisher={Frontiers Media SA},
   author={Meunier, David},
   year={2009} }

@article{dehghani2019computational,
  title={A computational perspective of the role of the thalamus in cognition},
  author={Dehghani, Nima and Wimmer, Ralf D},
  journal={Neural computation},
  volume={31},
  number={7},
  pages={1380--1418},
  year={2019},
  publisher={MIT Press One Rogers Street, Cambridge, MA 02142-1209, USA journals-info~…}
}

@article{zagha2020shaping,
  title={Shaping the cortical landscape: functions and mechanisms of top-down cortical feedback pathways},
  author={Zagha, Edward},
  journal={Frontiers in Systems Neuroscience},
  volume={14},
  pages={33},
  year={2020},
  publisher={Frontiers Media SA}
}

@article{nani2019neural,
  title={The neural correlates of consciousness and attention: two sister processes of the brain},
  author={Nani, Andrea and Manuello, Jordi and Mancuso, Lorenzo and Liloia, Donato and Costa, Tommaso and Cauda, Franco},
  journal={Frontiers in Neuroscience},
  volume={13},
  pages={1169},
  year={2019},
  publisher={Frontiers Media SA}
}

@inproceedings{thakare2023dyannet,
  title={Dyannet: A scene dynamicity guided self-trained video anomaly detection network},
  author={Thakare, Kamalakar Vijay and Raghuwanshi, Yash and Dogra, Debi Prosad and Choi, Heeseung and Kim, Ig-Jae},
  booktitle={Proceedings of the IEEE/CVF Winter conference on applications of computer vision},
  pages={5541--5550},
  year={2023}
}

@article{thakare2023rareanom,
  title={Rareanom: A benchmark video dataset for rare type anomalies},
  author={Thakare, Kamalakar Vijay and Dogra, Debi Prosad and Choi, Heeseung and Kim, Haksub and Kim, Ig-Jae},
  journal={Pattern Recognition},
  volume={140},
  pages={109567},
  year={2023},
  publisher={Elsevier}
}

@inproceedings{tur2023unsupervised,
  title={Unsupervised video anomaly detection with diffusion models conditioned on compact motion representations},
  author={Tur, Anil Osman and Dall’Asen, Nicola and Beyan, Cigdem and Ricci, Elisa},
  booktitle={International Conference on Image Analysis and Processing},
  pages={49--62},
  year={2023},
  organization={Springer}
}

@inproceedings{zaheer2022generative,
  title={Generative cooperative learning for unsupervised video anomaly detection},
  author={Zaheer, M Zaigham and Mahmood, Arif and Khan, M Haris and Segu, Mattia and Yu, Fisher and Lee, Seung-Ik},
  booktitle={Proceedings of the IEEE/CVF conference on computer vision and pattern recognition},
  pages={14744--14754},
  year={2022}
}

@inproceedings{guo2022deepcore,
  title={Deepcore: A comprehensive library for coreset selection in deep learning},
  author={Guo, Chengcheng and Zhao, Bo and Bai, Yanbing},
  booktitle={International Conference on Database and Expert Systems Applications},
  pages={181--195},
  year={2022},
  organization={Springer}
}

@inproceedings{yang2023towards,
  title={Towards sustainable learning: Coresets for data-efficient deep learning},
  author={Yang, Yu and Kang, Hao and Mirzasoleiman, Baharan},
  booktitle={International Conference on Machine Learning},
  pages={39314--39330},
  year={2023},
  organization={PMLR}
}

@misc{bai2025qwen25vltechnicalreport,
      title={Qwen2.5-VL Technical Report}, 
      author={Shuai Bai and Keqin Chen and Xuejing Liu and Jialin Wang and Wenbin Ge and Sibo Song and Kai Dang and Peng Wang and Shijie Wang and Jun Tang and Humen Zhong and Yuanzhi Zhu and Mingkun Yang and Zhaohai Li and Jianqiang Wan and Pengfei Wang and Wei Ding and Zheren Fu and Yiheng Xu and Jiabo Ye and Xi Zhang and Tianbao Xie and Zesen Cheng and Hang Zhang and Zhibo Yang and Haiyang Xu and Junyang Lin},
      year={2025},
      eprint={2502.13923},
      archivePrefix={arXiv},
      primaryClass={cs.CV},
      url={https://arxiv.org/abs/2502.13923}, 
}

@article{liu2024deepseek,
  title={Deepseek-v3 technical report},
  author={Liu, Aixin and Feng, Bei and Xue, Bing and Wang, Bingxuan and Wu, Bochao and Lu, Chengda and Zhao, Chenggang and Deng, Chengqi and Zhang, Chenyu and Ruan, Chong and others},
  journal={arXiv preprint arXiv:2412.19437},
  year={2024}
}
}
\newpage

\appendix

\section{Initialization and Formation of the Knowledge Prompt $\mathcal{P}$}
For the proposed framework, ReCoVAD, the knowledge prompt $\mathcal{P}$ plays a pivotal role in enabling effective knowledge transfer. Apart from our description of $\mathcal{P}$ in Section 3.2 of the main text, here we provide more information about its initialization and formation.

\begin{figure}[t]
    \centering
     \includegraphics[width=1.0\linewidth]
     {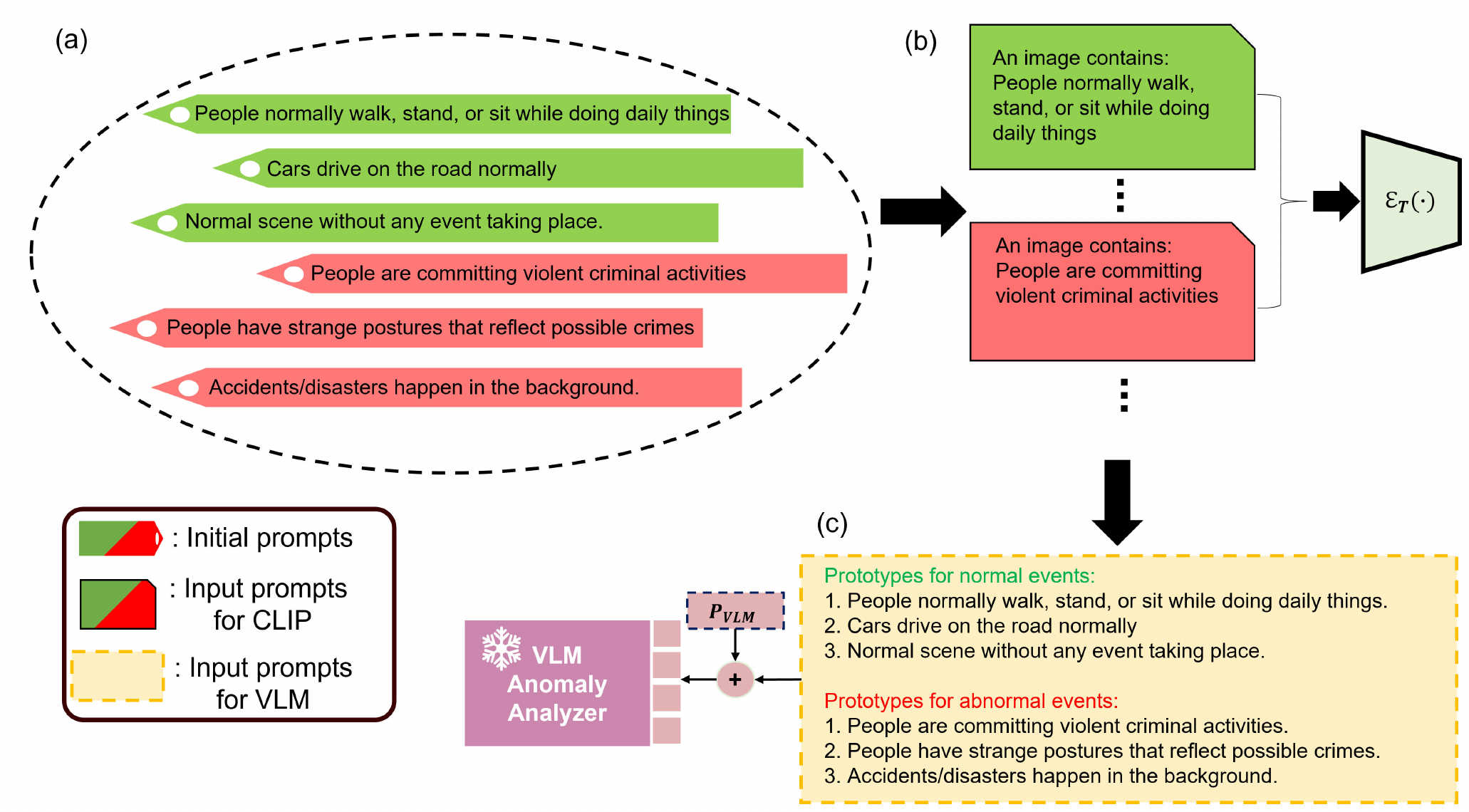}
     \caption{Initialization of the knowledge prompt $\mathcal{P}$.}
     \label{fig:inital}
\end{figure}

Although $\mathcal{P}$ is primarily updated in this work via the LLM reasoner in the conscious pathway by summarizing past cases, it is first initialized to provide the framework with essential prior knowledge. In this paper, we adopt the following unified prompt formulation to initialize $\mathcal{P}$ for both the UCF-Crime and XD-Violence datasets as in the Figure \ref{fig:inital} (a). The initial prompt list is evenly divided into the normal group and the abnormal group to maintain the information balance. The normal group is composed of three general descriptions of the normal event prototypes: \textit{1. People normally walk, stand, or sit while doing daily things. 2. Cars drive on the road normally. 3. Normal scene without any event taking place.} and three general descriptions of the abnormal event prototypes accordingly: \textit{People are committing violent criminal activities. 2. People have strange postures that reflect possible crimes. 3. Accidents/disasters happen in the background.}

These initial event prototypes provide a general description of both normal and abnormal events without delving into specific behaviors/actions, details that are instead summarized and extracted by the framework. We can easily formulate such descriptions without requiring intricate knowledge of the particular scenes or events involved. While maintaining their generality and simplicity, these prototypes also initialize the structure of the framework’s task space, offering a fundamental semantic partitioning between normal behaviors and abnormal events such as criminal activities and accidents. This, in turn, serves to guide subsequent large-scale models in the process of analyzing and summarizing the prototypes.

These prototypes are then rewritten as the knowledge prompt $\mathcal{P}$, a list of prompts formulated as \textit{"An image contains: {event prototype}}, as shown in \ref{fig:inital} (b), to enable the construction of the decision space in the reflex pathway.
The knowledge prototypes in the prompt list above are concatenated in the form of the template in Figure \ref{fig:inital} (c), as a code book for the VLM anomaly analyzer to refer to and make their descriptions and judgments of whether the image contains abnormal events. Finally, we require the LLM reasoner to output $L$ prototypes from the previous cases, following the same form as the initial descriptions. Also, to maintain the normal/abnormal balance in the prompts, we ask the LLM to output up to $L/2$ normal prototypes and $L/2$ abnormal prototypes to make sure the new abnormal knowledge is recorded in the sample-unbalanced anomaly detection task.

\section{Formulation of the Prompt $P_{VLM}$}
\begin{figure}[t]
    \centering
     \includegraphics[width=1.0\linewidth]
     {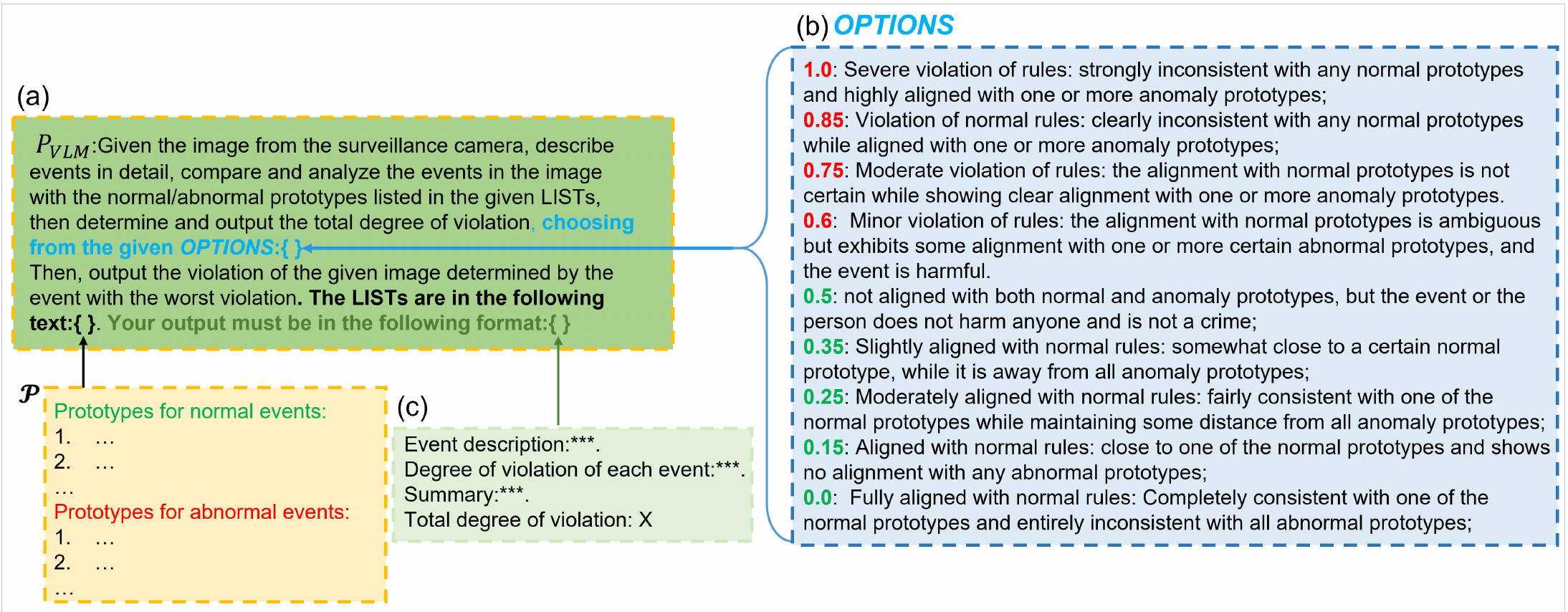}
     \caption{The design of the $P_{VLM}$ prompt for the VLM anomaly analyzer.}
     \label{fig:vlm}
\end{figure}

In this section, we discuss the formation of the prompt $P_{VLM}$. As described in Section 3.4, the prompt fulfills the following functions: (1) describe events in the frame; (2) compare them with
prototypes in the knowledge prompt $\mathcal{P}$; (3) assign an anomaly score chosen from the option list: \textit{OPTIONS}; (4) regulate the output format for future analysis. In order to instruct the VLM anomaly analyzer to achieve these functions, we design the $P_{VLM}$ as in Figure \ref{fig:vlm}.

As is shown in the Figure \ref{fig:vlm} (a), the overall formation of the prompt $P_{VLM}$ instructs the VLM anomaly analyzer to describe all the events in the image in detail, compare them against the event prototypes in the given knowledge prompt $\mathcal{P}$( in the code book form) and determine their anomaly score, which are chosen from the option list \textit{OPTION} in the Figure \ref{fig:vlm} (b).

As demonstrated, \textit{OPTION} constrains the scores the model can choose from and requires it to make decisions based on explicit interpretations of these scores, thereby reducing the possibility of subjective or hasty judgments. Each score explanation is derived from the degree of alignment between the observed event and the predefined prototypes of normal or abnormal events. The scores are determined based on the following principle: if an event shows alignment with one or more abnormal prototypes, it should be assigned a score higher than 0.5, with the exact score reflecting the strength of this alignment. Conversely, if the event does not match any abnormal prototype, it is assigned a low anomaly score, which is determined based on how well it aligns with normal prototypes; the stronger the alignment with normal behavior, the lower the 
anomaly score.

Further, $P_{VLM}$ asks the model to determine the degree of the event of the worst anomaly situation, as the image may contain normal events that can hinder the model's decision. At last, the VLM anomaly analyzer is required to output the information in the given format in the Figure \ref{fig:vlm} (c) to allow the extraction of the event descriptions and anomaly scores for the LLM and the reflex pathway. 

Notably, in the main text, we only demonstrate the \textbf{Summary} and \textbf{Total degree of violation} part of the VLM output for visualization in Figure 3 due to the length limit.

\section{Formulation of the Prompt $P_{LLM}$}

\begin{figure}[t]
    \centering
     \includegraphics[width=1.0\linewidth]
     {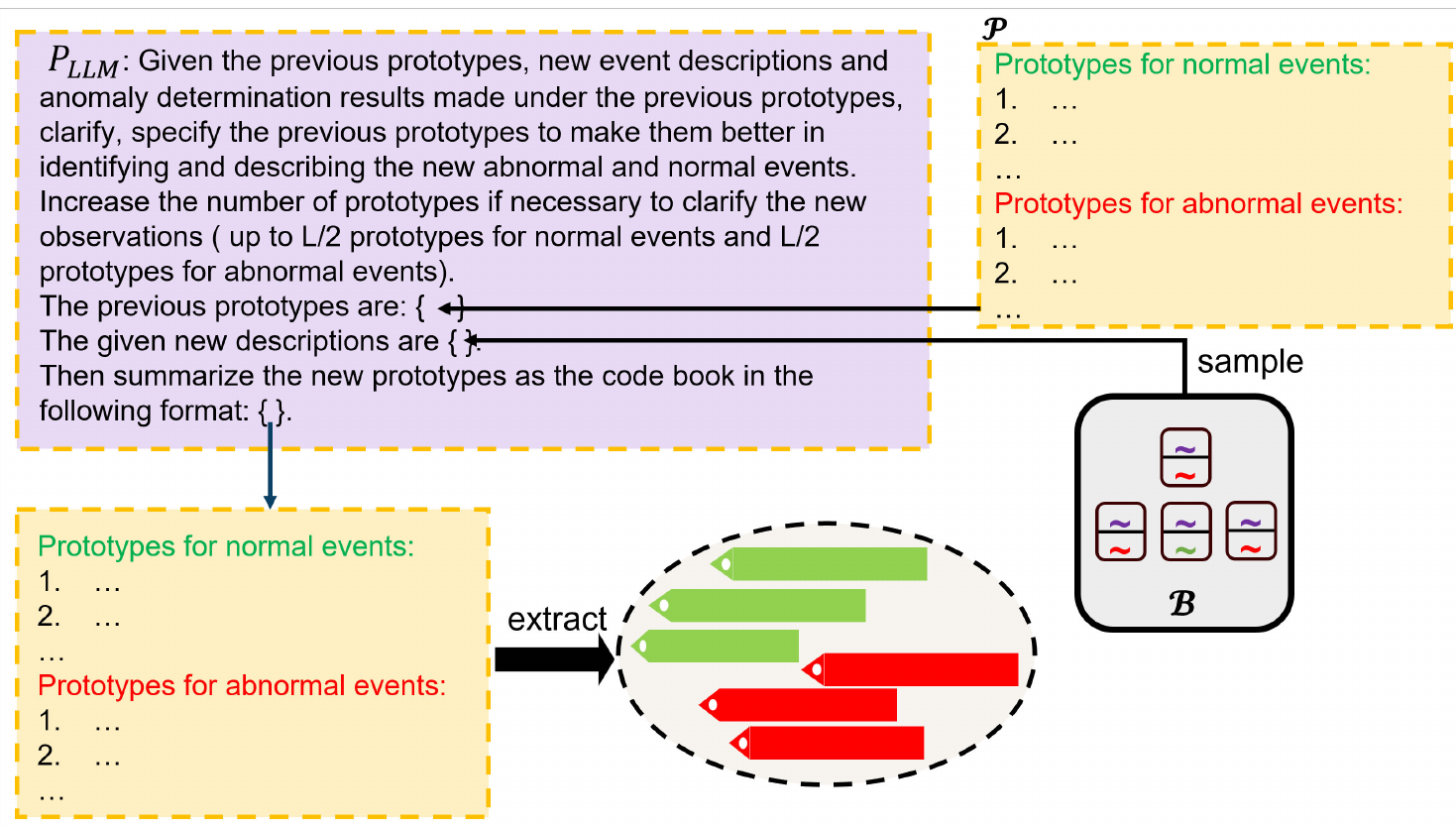}
     \caption{The design of the $P_{LLM}$ prompt for the LLM reasoner.}
     \label{fig:llm}
\end{figure}

In this section, we further introduce the prompt $P_{LLM}$ used in the conscious pathway. The primary function of this prompt is to instruct the model to leverage the accumulated historical anomaly descriptions/detection cases to revise and expand the previous knowledge prompt $\mathcal{P}$, finally generating an updated version. The detailed information about the prompt $P_{LLM}$ is shown in Figure \ref{fig:llm}. 

As is demonstrated, $P_{LLM}$ gathers the previous cases sampled from the description set $\mathcal{B}$ and instructs the LLM to clarify and specify the previous prompts to fit the new cases. This process requires the model to analyze previously used event prototypes that may have led to inaccurate discrimination, and to concretize vague event descriptions by incorporating new event information. In doing so, the updated knowledge prompt can better support the lower-layer model in sample filtering and anomaly event discrimination. The outputs are new prototypes which are organized in the same form as the code book form of  $\mathcal{P}$. These prototypes are extracted from the output and are treated as the new knowledge prompt, fed back to the reflex pathway and the VLM anomaly analyzer.

\section{Ablations on the Parameter $N$ and the Parameter $L$}

In this section, we present additional ablation studies to further validate our framework's effectiveness. Specifically, the ablations focus on two key parameters: the interval parameter $N$, which determines the frequency of LLM reasoning, and the event prototype count parameter $L$, which controls the number of event prototypes maintained in the knowledge prompt $\mathcal{P}$.

\begin{figure}[t]
    \centering
     \includegraphics[width=1.0\linewidth]
     {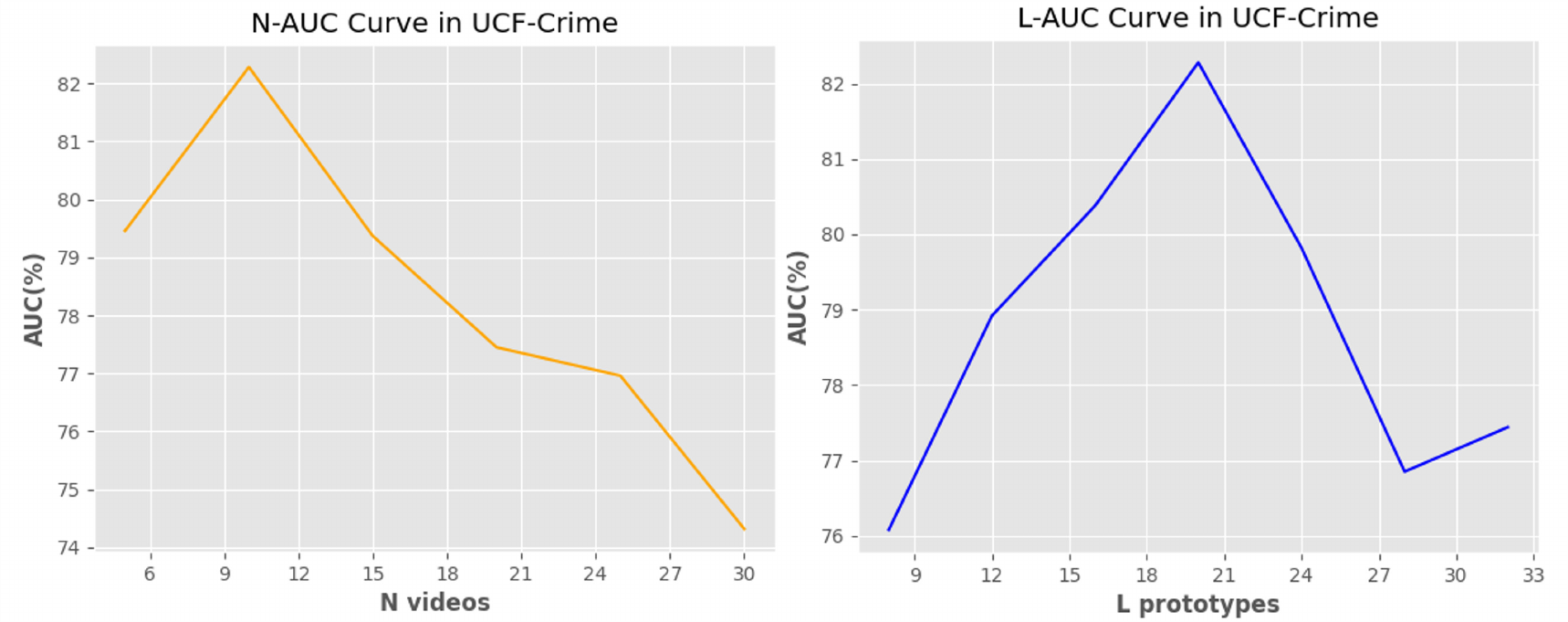}
     \caption{The ablation on the parameter $N$ and the parameter $L$.}
     \label{fig:nl}
\end{figure}

The parameter $N$ is a critical component of our framework, as it determines how frequently the LLM is invoked and, consequently, how often the memory $\mathcal{M}$ and detection results are refined. In practice, we define $N$ in terms of the total number of frames corresponding to $N$ videos for convenience. 

As shown in Figure \ref{fig:nl}, we report the results under different values of $N$. It is evident that overly frequent summarization does not improve model performance; in fact, it leads to a noticeable degradation. This can be attributed to the limited scope of prototype summarization when the LLM is invoked too frequently, which weakens the generalization ability of the event prototypes. Frequent updates to the decision space also introduce instability, further degrading performance.

Conversely, setting $N$ too large also results in significant performance drops. In such cases, as shown in Figure \ref{fig:nl}, the large volume of videos causes an accumulation of event descriptions, making novel and informative anomalies relatively sparse. As a result, the model struggles to extract meaningful new knowledge. Moreover, a long summarization interval delays the update of outdated

In addition, we evaluate the impact of the prototype count parameter $L$ on the performance of our framework. The results are presented in Figure \ref{fig:nl}. As shown, when $L$ is too small, the upper bound of knowledge that the framework can retain is limited, making it insufficient to capture the complexity of the dataset, especially for datasets like UCF-Crime and XD-Violence, which exhibit highly diverse and complex distributions. This leads to a drop in detection accuracy.

On the other hand, when $L$ is set too large, the knowledge prompt $\mathcal{P}$ becomes overly overpopulated with redundant information, which hinders the clear separation between different types of events within the decision space. Moreover, the excessive information may introduce unnecessary interference for the VLM anomaly analyzer when distinguishing between normal and abnormal events, ultimately resulting in degraded performance.

\section{Visualization of ReCoVAD's Self-evolution Via the Knowledge Prompt $\mathcal{P}$}

In this section, we demonstrate the self-evolution capability of our framework, ReCoVAD, through the visualization of knowledge prompt $\mathcal{P}$'s self-refinement. As discussed earlier, $\mathcal{P}$ is a crucial component of our framework, serving as the bridge between the reflex and conscious pathways. They provide essential guidance to the VLM anomaly analyzer and play a key role in shaping the decision space of the reflex pathway. More specific and task-aligned knowledge prompt leads to better overall performance of the framework in video anomaly detection.

\begin{figure}[t]
    \centering
     \includegraphics[width=1.0\linewidth]
     {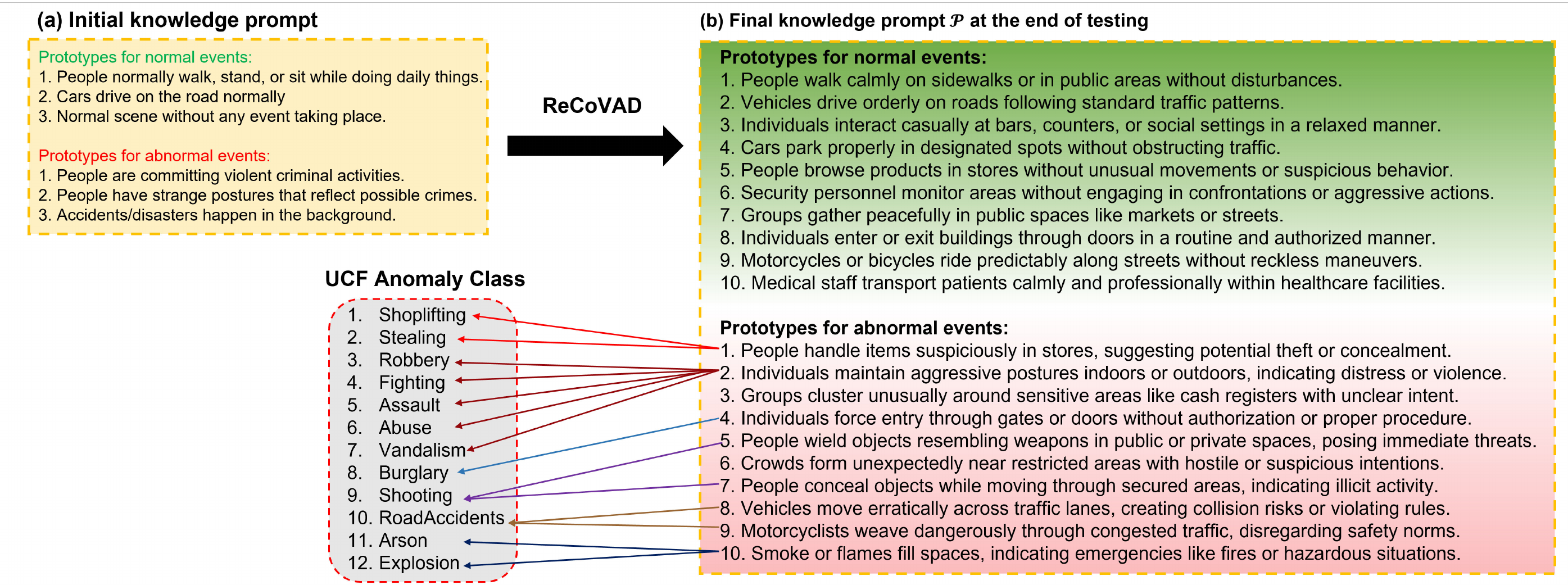}
     \caption{Visualization of the knowledge prompt $\mathcal{P}$ at the end of the testing on the UCF-Crime.}
     \label{fig:prompt_ucf}
\end{figure}

\begin{figure}[t]
    \centering
     \includegraphics[width=1.0\linewidth]
     {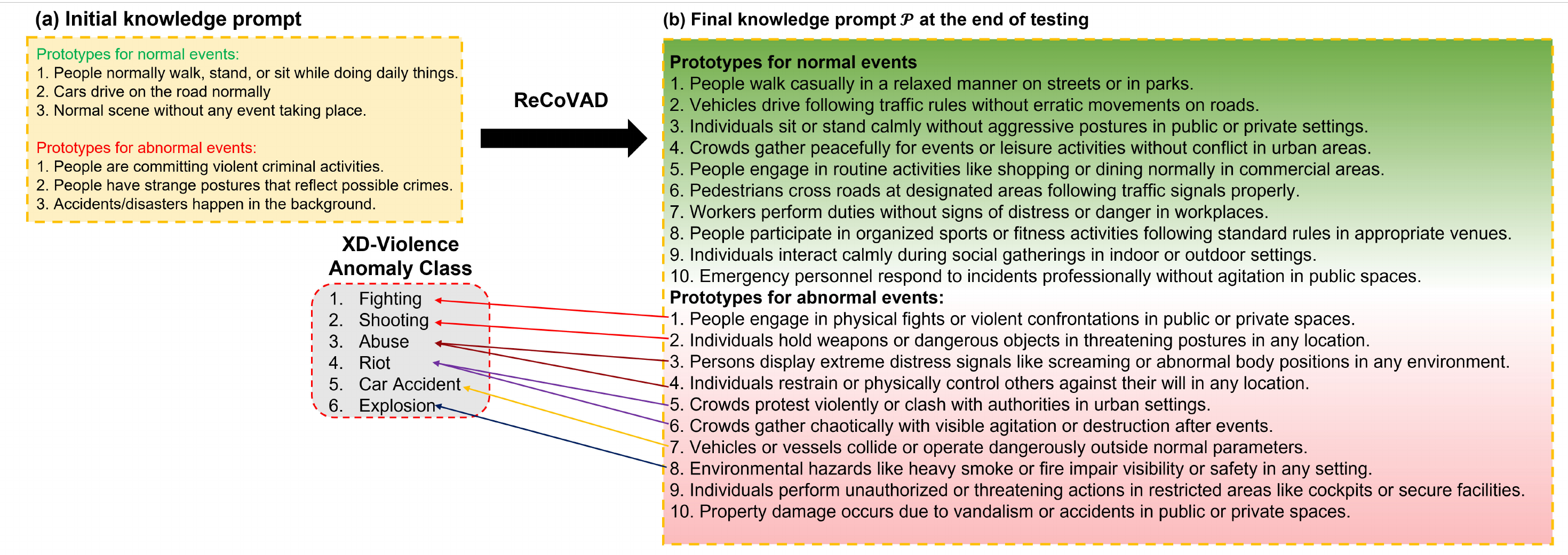}
     \caption{Visualization of the knowledge prompt $\mathcal{P}$ at the end of the testing on the XD-Violence.}
     \label{fig:prompt_xd}
\end{figure}

Figure \ref{fig:prompt_ucf} and Figure \ref{fig:prompt_xd} illustrate the final knowledge prompt obtained after testing. For the sake of visual clarity, we formulate the knowledge prompt in the code book form. Remarkably, without any human intervention or supervision during the process, our framework can automatically extract and summarize the types of abnormal events present in the dataset through a combination of top-down refinement and bottom-up filtering. As indicated by the arrows connecting the abnormal event prototypes to the ground-truth anomaly classes, most of the event prototypes in the knowledge prompt $\mathcal{P}$ exhibit a clear semantic alignment with one or more ground-truth anomaly categories. This demonstrates that our framework can effectively identify key information about abnormal events through its self-evolving process. As a result, it is capable of autonomously adapting to different datasets, significantly enriching and grounding the initial knowledge prompt, and providing meaningful guidance for the models in the lower layers.

It is important to note that the initial prompts did not contain any dataset-specific information or annotations regarding the types of abnormal events. This highlights the framework’s strong capacity for self-evolution and knowledge discovery.

\section{Broader Impacts}
Our proposed ReCoVAD framework introduces a highly efficient, training-free video anomaly detection approach that significantly reduces computational burden while achieving state-of-the-art performance. By processing only a fraction of video frames with large pretrained models, ReCoVAD enables real-time deployment in resource-constrained environments such as edge devices for surveillance, industrial monitoring, and autonomous systems. This can lead to broader accessibility of intelligent safety systems in low-resource regions and energy-efficient AI solutions. Furthermore, the dual-pathway design inspired by human cognition, which integrates reflex arcs and conscious reasoning, opens a new direction for adaptive, cost-aware inference in general, large-model-based systems. The ability to detect novel anomalies using language-guided memory refinement also promotes robustness in dynamic, real-world environments.

\end{document}